\documentclass[letterpaper]{article} 
\usepackage{aaai25}  
\usepackage{times}  
\usepackage{helvet}  
\usepackage{courier}  
\usepackage[hyphens]{url}  
\usepackage{graphicx} 
\usepackage{natbib}  
\usepackage{caption} 
\usepackage{algorithm}
\usepackage{algorithmic}

\usepackage{newfloat}
\usepackage{listings}
\DeclareCaptionStyle{ruled}{labelfont=normalfont,labelsep=colon,strut=off} 
\floatstyle{ruled}
\newfloat{listing}{tb}{lst}{}
\floatname{listing}{Listing}
\usepackage{multirow}
\usepackage{subfigure}
\usepackage{subcaption}
\usepackage{amsmath}
\usepackage{amsfonts}

\title{Diverse Rare Sample Generation with Pretrained GANs}
\author{
    Subeen Lee\textsuperscript{\rm 1},
    Jiyeon Han\textsuperscript{\rm 1},
    Soyeon Kim\textsuperscript{\rm 1} and
    Jaesik Choi\textsuperscript{\rm 1,2}
}
\affiliations{
    \textsuperscript{\rm 1}Korea Advanced Institute of Science and Technology (KAIST), South Korea\\
    \textsuperscript{\rm 2}INEEJI, South Korea\\
    \{sbrblee7, j.han, soyeon.k, jaesik.choi\}@kaist.ac.kr
}

\begin{document}

\maketitle

\begin{abstract}
Deep generative models are proficient in generating realistic data but struggle with producing rare samples in low density regions due to their scarcity of training datasets and the mode collapse problem.
While recent methods aim to improve the fidelity of generated samples, they often reduce diversity and coverage by ignoring rare and novel samples. This study proposes a novel approach for generating diverse rare samples from high-resolution image datasets with pretrained GANs. Our method employs gradient-based optimization of latent vectors within a multi-objective framework and utilizes normalizing flows for density estimation on the feature space. This enables the generation of diverse rare images, with controllable parameters for rarity, diversity, and similarity to a reference image. We demonstrate the effectiveness of our approach both qualitatively and quantitatively across various datasets and GANs without retraining or fine-tuning the pretrained GANs.
\end{abstract}

%
\begin{links}
    \link{Code}{https://github.com/sbrblee/DivRareGen}
\end{links}

\section{Introduction}

Deep generative models have shown impressive generative capabilities across various domains.
The primary focus of current generative model research is on enhancing the fidelity of generated images \cite{devries2020instance, karras2019style, azadi2018discriminator, turner2019metropolis}. However, these approaches often compromise sample diversity and encounter difficulties generating rare samples, primarily due to their limited representation in the training dataset \cite{sehwag2022generating, lee2021self}. In GANs, this issue is worsened by the mode collapse problem \cite{thanh2020catastrophic}. Investigating rare samples is crucial for several reasons: it enhances the creation of synthetic datasets that embody diversity and creativity \cite{sehwag2022generating, agarwal2022estimating}, ensures fairness in generative processes \cite{teo2023fair, hwang2020fairfacegan}, and aligns with the human tendency to favor unique features \cite{snyder2001handbook,lynn1997individual}. Additionally, exploring edge cases and unusual scenarios is essential in various domains, such as drug discovery or molecular design \cite{sagar2023physics, zeng2022deep}, and natural hazard analysis \cite{ma2024generative}.

Several studies have been conducted to enhance the overall diversity of GAN-generated outputs and to promote the generation of rare samples \cite{chang2024quality, allahyani2023divgan, humayun2022polarity, humayun2021magnet, heyrani2021creativegan, ghosh2018multi, tolstikhin2017adagan, srivastava2017veegan, chen2016infogan}. Due to the high computational cost of training GANs \cite{karras2019style, brock2018large}, techniques without model retraining are appealing. For example, \citet{humayun2022polarity} proposed a resampling technique for pretrained GANs with a controllable fidelity-diversity tradeoff parameter. However, the proposed method requires extensive sampling to cover the data manifold fully. On the other hand, \citet{chang2024quality} proposed a method to obtain diverse samples that satisfy text conditions by optimizing latent vectors with a quality-diversity objective.

\citet{han2023rarity} proposes a rarity score for samples using relative density measures based on $k$-nearest neighbor ($k$-NN) manifolds in the feature space of pretrained classifiers. Although $k$-NN density estimation is straightforward and reliable \cite{naeem2020reliable,kynkaanniemi2019improved}, its non-differentiable nature complicates gradient-based optimization. In contrast, normalizing flows (NFs) excel at high-dimensional density estimation in differentiable form \cite{papamakarios2021normalizing, kingma2018glow, dinh2016density, dinh2014nice}. We employ NFs for density estimation in the feature space, which incurs a lower training cost compared to retraining or fine-tuning GANs, and analyze how NF-based density estimates relate to the rarity score.

\begin{figure*}
    \centering
    \includegraphics[width=0.99\linewidth]{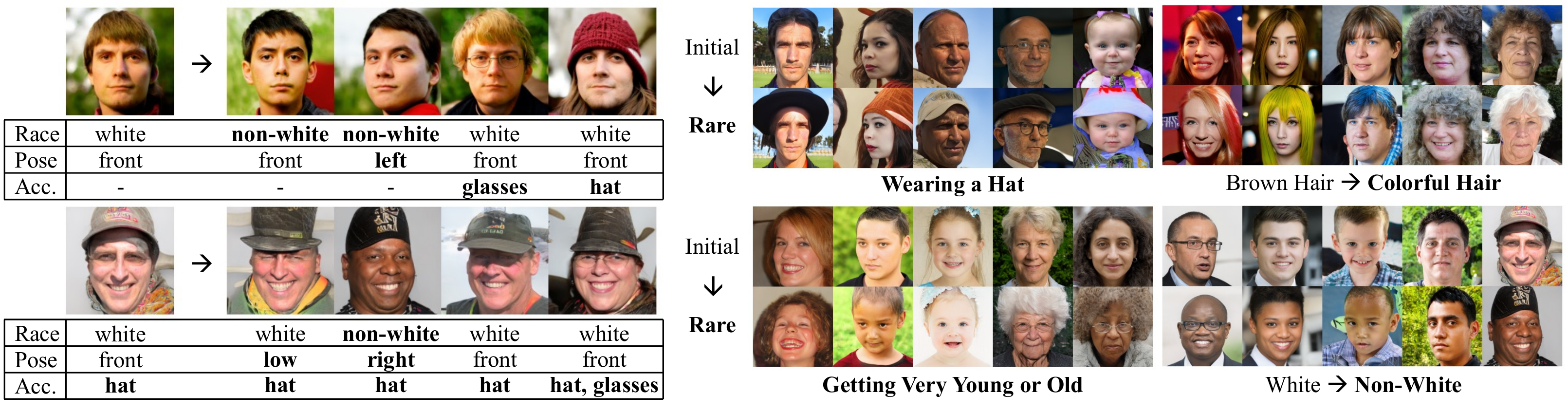}
    \caption{Examples of rare samples generated by our method. \textbf{Left}: Our method produces diverse rare images for a single reference, with variations even within the same rare attribute (e.g., hats of different shapes and colors). \textbf{Right}: Generated rare attributes include accessories like hats, non-brown hair colors, extreme ages, and non-white races. ``Pose'' refers to head orientation, and ``Acc.'' denotes accessories. Rare attributes are highlighted in bold.}
    \label{fig:1}
\end{figure*}

This study aims to generate diverse rare samples for a given high-resolution image datasets and GANs. Our method does not require any fine-tuning or retraining of the GANs but instead explores the latent space of the given model through gradient-based optimization. Our contributions are as follows:
\begin{itemize}
\item Our method can generate diverse versions of rare images utilizing the multi-start method in optimization, without being trapped at the same local optima.
\item Rarity and diversity of generated images and similarity to the initial image can be controlled via a multi-objective optimization framework.
\item We demonstrate the effectiveness of our method with various high-resolution image datasets and GANs, both qualitatively and quantitatively.
\end{itemize}

\section{Related Work}
\paragraph{Rare Generation}
\citet{han2023rarity} introduced the rarity score to quantify the uniqueness of individual samples, distinguishing it from conventional metrics that primarily evaluate fidelity or diversity in generated samples \cite{kynkaanniemi2019improved, zhang2018unreasonableeffectivenessdeepfeatures, heusel2017gans}. The rarity score is defined as the minimum $k$-nearest neighbor distance ($k$-NND) among real samples that are closer to the target sample, with higher scores indicating lower density within the real data manifold. However, obtaining rare samples has received limited attention. \citet{sehwag2022generating} addressed this by leveraging a pretrained classifier to estimate likelihoods and adapting the sampling process of diffusion probabilistic models to target low-density regions while maintaining fidelity. Their method focuses on sampling from regions far from class mean vectors in the feature space and penalizes deviations from the overall mean vector of real data. However, this approach is class-conditional and depends on a Gaussian likelihood function.

On the other hand, \citet{humayun2022polarity} tackled the mode collapse issue in GANs with Polarity sampling, a fidelity-diversity controllable resampling strategy for pretrained GANs. It approximates the GAN's output space using continuous piecewise affine splines. By tuning $\rho$, sampling can focus on modes ($\rho<0$) or anti-modes ($\rho>0$), with higher $\rho$ increasing diversity by targeting low-density regions. However, it does not guarantee the fidelity of the selected samples and requires extensive sampling and Jacobian matrix computations, leading to high computational costs.

\paragraph{Quality-Preserved Diverse Generation Using Pretrained GANs}
Generating rare samples is important, but maintaining quality is also crucial for their usefulness \cite{amabile2018creativity}. Achieving diverse, high-fidelity samples is similar to finding multiple solutions in combinatorial optimization. \citet{chang2024quality} addressed this by proposing a quality-diversity algorithm that updates the latent vector to balance quality and diversity, using the CLIP score \cite{radford2021learning} to measure similarity and diversity.
In our work, we also optimize the latent vector to generate diverse samples; however, we prioritize rarity as the main objective rather than quality and use Euclidean distance in arbitrary feature spaces, avoiding the additional text constraints required by the CLIP score.
To prevent low-fidelity samples, we apply a constraint to keep the sampled data within the real data manifold.

\paragraph{Reference-based Generation}
Finding diverse rare variations of a given initial image relates to reference-based image generation, encompassing tasks like domain adaptation \cite{yang2023one}, editing \cite{xia2023feditnet}, and conditional generation \cite{casanova2021instance}. While these approaches involve additional training costs for each attribute \cite{yang2023one} or reference \cite{xia2023feditnet}, or require a different GAN training scheme \cite{casanova2021instance}, our method only requires a single training phase for the density estimator across multiple references with pretrained GANs.

\begin{figure*}
    \centering
    \includegraphics[width=0.99\linewidth]{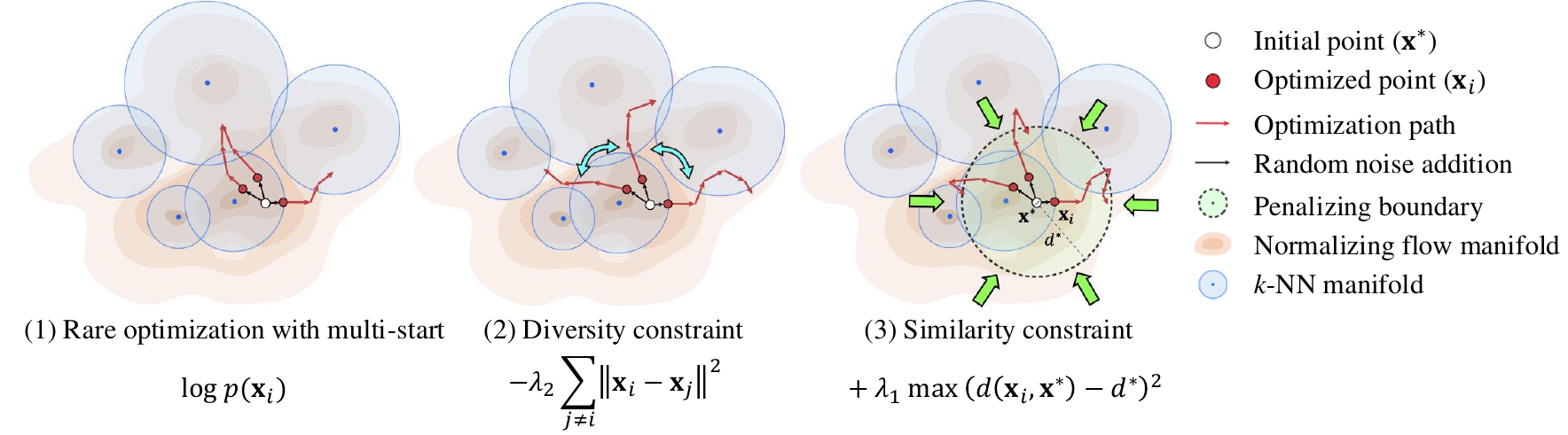}
    \caption{Schematic diagram for the objective function of our method. $\mathbf{x}^*=f(G(\mathbf{z}^*))$ and $\mathbf{x}_i=f(G(\mathbf{z}_i))$ for brevity.}
    \label{fig:2}
\end{figure*}

\paragraph{Density Estimation for Images}
The rarity score identifies samples in low-density regions of the real data manifold, making it valuable for detecting rare generations. However, the non-differentiable nature of $k$-NN-based manifold estimation complicates gradient-based optimization for directly obtaining rare samples. In contrast, extensive research has been conducted to estimate density in high-dimensional spaces. Normalizing flows (NFs), as likelihood-based probabilistic models, use a sequence of invertible functions to transform a simple density into a complex one, potentially representing multi-modal distributions while preserving data relationships \cite{papamakarios2021normalizing, kingma2018glow, dinh2016density, dinh2014nice}. 
While NFs may struggle with out-of-distribution data due to their focus on low-level features \cite{kirichenko2020normalizing}, training them on the feature space of a pretrained network—which includes high-level semantic information—can help mitigate this issue \cite{esser2020disentangling}.

\paragraph{Multi-Start Method for Diverse Solutions}
We frame our problem of obtaining diverse rare samples for each given reference as identifying multiple local minima around the reference in the data distribution. A straightforward approach to this problem is the multi-start method, an algorithm that iteratively searches for local optima starting from multiple initial points \cite{feo1995greedy, rochat1995probabilistic, rinnooy1987stochastic}. This method selects several starting positions and applies a local search algorithm to each, aiming to locate distinct local optima. Although easy to implement, it does not always ensure that different starting points result in different local optima \cite{tarek2022simplifying}. To address this issue, we add the diversity and similarity constraints to the objective function, ensuring that each initial point converges to a different minimum.

\section{Methods}
\subsection{Problem Statement}
Given a GAN generator $G=G(\mathbf{z})$, an arbitrary initial latent vector $\mathbf{z}^*\in\mathbb{R}^m$, a feature extractor $f=f(\mathbf{I})$, and a density estimator $g=g(\mathbf{x})$, our objective is to find a set of latent vectors $\{\mathbf{z}_i\}^N_{i=1}$ that generates diverse rare samples which are similar to the image generated from $\mathbf{z}^*$ (referred to as the reference for the rest of the paper). Here, $\mathbf{I}\in\mathbb{R}^{w\times h \times 3}$ and $\mathbf{x}\in\mathbb{R}^n$ denote an image and a feature vector, respectively. For simplicity, we denote $\mathbf{x}^*=f(G(\mathbf{z}^*))$ and $\mathbf{x}_i=f(G(\mathbf{z}_i))$. In this study, we define similarity as the Euclidean distance in the feature space, $d(\mathbf{x}_1,\mathbf{x}_2)=\|\mathbf{x}_1-\mathbf{x}_2\|$, a metric shown to align well with human perception \cite{zhang2018unreasonableeffectivenessdeepfeatures}.

We propose a multi-objective optimization framework that integrates rarity, diversity, and similarity regularization, as illustrated in Fig.~\ref{fig:2}, and provide a detailed explanation in the subsequent sections.

\subsection{Rare Sample Generation}
For rarity, the density estimated from $g$ is used. NFs are employed due to their remarkable performance in high-dimensional density estimation, though any differentiable density estimator can be applied. NFs provide the exact log-likelihood of individual samples, allowing us to directly define the objective function as $\mathcal{L}_{rare}(\mathbf{x})=g(\mathbf{x})=\log p(\mathbf{x})$ to minimize. To control the similarity between the generated rare image and the reference image, we incorporate a regularization term inspired by \citet{chang2024quality}. Specifically, we define the similarity loss as $\mathcal{L}_{sim}(\mathbf{x})=(\max (d(\mathbf{x}, \mathbf{x}^*), d^*)-d^*)^2$, where this term penalizes samples that exceed a predefined boundary, referred to as the \textit{penalizing boundary} throughout this paper, defined by $d^*$. Specifically, we use the distance to $k'$-nearest neighbor in the fake $k$-NN manifold for $d^*$. We also strictly accept the sample inside this boundary as well as inside the real $k$-NN manifold $\Phi_{real}$ in optimization. The optimization goal is formulated by combining the two objectives—rarity and similarity regularization—as follows.
\begin{equation}
\begin{aligned}
    \min_{\substack{\mathbf{z}\\ \mathbf{x}=f(G(\mathbf{z}))}} \mathcal{L}_{rare}(\mathbf{x}) 
    + \lambda_1 \mathcal{L}_{sim}(\mathbf{x})\\
    \textrm{subject to $\mathbf{x}\in\Phi_{real}$ and $d(\mathbf{x}, \mathbf{x}^*)\leq d^*$}
\end{aligned}
\end{equation}
\subsection{Diverse Rare Sample Generation}
We utilize the multi-start method to obtain diverse rare images by adding small random noises to $\mathbf{z}^*$, generating multiple starting points for optimization. Specifically, $\{\mathbf{z}_i\}^N_{i=1}$ are initialized as $\mathbf{z}_i=\mathbf{z}^*+\epsilon$, where $\epsilon\sim\mathcal{N}(\mathbf{0}, \sigma^2I)$. However, as shown in Fig.~\ref{fig:2} (1), this does not guarantee obtaining different rare images, but they might converge to the same local minima. To address this issue, we add a diversity constraint to the objective, $\mathcal{L}_{div}(\mathbf{x}_i)=-\sum_{j\neq i} d(\mathbf{x}_i, \mathbf{x}_j)^2$, ensuring that the samples are far from each other, as shown in Fig.~\ref{fig:2} (2). This term is inspired by the expected distances in feature space, similar to the concept of Maximum Mean Discrepancy (MMD) \cite{gretton2006kernel}. Combining all objectives, the multi-objective optimization problem is formulated as follows.
\begin{equation}
\hfill
\begin{aligned}
    \min_{\substack{\mathbf{z}_i\\ \mathbf{x}_i=f(G(\mathbf{z}_i))}} \mathcal{L}_{rare}(\mathbf{x}_i) 
    + \lambda_1 \mathcal{L}_{sim}(\mathbf{x}_i) + \lambda_2 \mathcal{L}_{div}(\mathbf{x}_i)  \\
    \textrm{subject to $\mathbf{x}_i\in\Phi_{real}$ and $d(\mathbf{x}_i, \mathbf{x}^*)\leq d^*$}
\end{aligned}
\label{eq:diverse}
\end{equation}

\section{Experimental Results}
We validate our proposed method using high-resolution image datasets with a resolution of $1024\times1024$, including Flickr Faces HQ (FFHQ) \cite{karras2019style}, Animal Faces HQ (AFHQ) \cite{choi2020stargan}, and Metfaces \cite{karras2020training}. StyleGAN2 with config-f \cite{karras2020analyzing} and StyleGAN2-ADA \cite{karras2020training} are utilized. For feature extraction, we employ the VGG16-fc2 architecture \cite{simonyanZ14a}. As the density estimator, the Glow architecture \cite{kingma2018glow} is adapted to accommodate the high dimensionality of the feature space. The optimization is performed using the Adam optimizer \cite{diederik2014adam} with a learning rate of $2\times 10^{-2}$, combined with a StepLR scheduler. The best optimization results are recorded when the lowest loss is achieved according to Equation (\ref{eq:diverse}). Additional details including computational cost are provided in Appendix B.

\subsection{Generation of Rare Facial Attributes with StyleGAN2 (FFHQ-StyleGAN2)}
\subsubsection{Quantitative Results}
As a baseline, 10,000 synthetic samples are generated using latent vectors from StyleGAN2 with a truncation parameter of $\psi=1.0$. With our method, we generate ten rare samples for each of 1,000 initial latent vectors from the baselines, using parameters $\lambda_1=30.0, \lambda_2=0.002, \sigma=0.1$, and $k'=100$\footnote{For the fake $k$-NN manifold estimation, 10,000 generated samples are used.}. The choice of the parameters is explained in Appendix C. For Polarity sampling, 250,000 latent vectors and their corresponding Jacobian matrices are obtained from the authors \cite{humayun2022polarity}, and 10,000 samples are resampled using $\rho=[1.0,5.0]$ (anti-mode sampling).

Results are evaluated with metrics including the Rarity Score (RS) \cite{han2023rarity}, precision (Prec.) and recall (Rec.) for fidelity and diversity \cite{kynkaanniemi2019improved}, LPIPS score for diversity~\cite{zhang2018unreasonableeffectivenessdeepfeatures}, and FID score \cite{heusel2017gans}, as shown in Table~\ref{tab:3}. Each metric is computed using 10,000 generated samples, with the LPIPS score averaged over 10,000 random sample pairs. Significant differences in LPIPS scores between sampling methods are confirmed using an unpaired t-test. For the real $k$-NN manifold, $k=3$ is used.

Our method improves both rarity and diversity compared to the baseline, even when the optimization uses only 10\% of the baseline samples as references. The FID score decreases because more samples are generated in low-density regions, reducing samples near the data distribution's modes. Polarity sampling also enhances rarity and diversity but sacrifices precision, as it primarily targets low-density regions in the GAN's output space rather than the real manifold, often generating out-of-distribution samples (structural zeros; \cite{kim2023deep}). Furthermore, in contrast to our objective of generating rare samples similar to a given reference, Polarity sampling is not designed for it. Finally, since Polarity sampling operates by resampling from an initial set, its diversity is heavily dependent on the size of that initial set. An additional comparison with Polarity sampling is in Appendix G.

\setlength{\tabcolsep}{0.6mm}
\begin{table}[!t]
\centering
\begin{tabular}{l|rrrrr}
\hline
\textbf{\,Model} & \textbf{RS} $\uparrow$ & \textbf{Prec.} $\uparrow$ & \textbf{Rec.} $\uparrow$ & \textbf{LPIPS} $\uparrow$ & \textbf{FID} $\downarrow$ \\ \hline
\textbf{\,Baseline} & 18.88 & 0.69 & 0.56 & 0.73 & \textbf{4.17} \\ 
\textbf{\,Polarity $(\rho=1.0)$} & 24.71 & 0.39 & 0.70 & 0.75 & 33.28 \\ 
\textbf{\,Polarity $(\rho=5.0)$} & \textbf{24.83} & 0.38 & \textbf{0.71} & 0.75 & 34.11 \\ 
\textbf{\,Ours} & 23.50 & \textbf{0.92} & 0.65 & \textbf{0.76} & 7.38 \\ \hline
\end{tabular}
\caption{Quantitative evaluation for Section 4.1.}
\label{tab:3}
\end{table}

\setlength{\tabcolsep}{1.3mm}
\begin{table}[!t]
\begin{tabular}{ll|rrr}
\hline
\multicolumn{2}{l|}{\textbf{Model-Based Attribute}} & \textbf{FFHQ} & \textbf{Reference} & \textbf{Ours(\%)} \\ \hline
\multicolumn{1}{l}{} & \textit{\textbf{0-9}} & 7.51 & 7.60 & \textbf{8.56}      \\ 
\multicolumn{1}{l}{} & \textit{\textbf{10-69}}& 92.23& \textbf{92.20} & 90.98\\ 
\multicolumn{1}{l}{\multirow{-3}{*}{\textbf{Age}}} & \textit{\textbf{Over70}} & 0.25 & 0.20 & \textbf{0.45}      \\ \hline
\multicolumn{1}{l}{} & \textit{\textbf{Male}} & 45.46& 47.44 & \textbf{48.50}     \\ 
\multicolumn{1}{l}{\multirow{-2}{*}{\textbf{Gender}}} & \textit{\textbf{Female}} & 54.53& \textbf{52.55} & 51.49\\ \hline
\multicolumn{1}{l}{} & \textit{\textbf{White}}& 62.97& \textbf{64.66} & 61.52\\ 
\multicolumn{1}{l}{\multirow{-2}{*}{\textbf{Race}}} & \textit{\textbf{NonWhite}} & 37.03& 35.33 & \textbf{38.47}     \\ \hline
\multicolumn{1}{l}{} & \textit{\textbf{Front}}& 60.98& \textbf{55.25} & 45.73\\ 
\multicolumn{1}{l}{\multirow{-2}{*}{\textbf{HeadPose}}} & \textit{\textbf{NotFront}} & 39.01& 44.75 & \textbf{54.26}     \\ 
\hline
\end{tabular}
\caption{Percentage of age, gender, race, and head pose attributes predicted by FaceXformer for Section 4.1.}
\label{tab:1}
\end{table}

\setlength{\tabcolsep}{1.2mm}
\begin{table}[!t]
\centering
\begin{tabular}{ll|rrr}
\hline
\multicolumn{2}{l|}{\textbf{LFWA Attribute}} & \textbf{FFHQ} & \textbf{Reference} & \textbf{Ours(\%)} \\ \hline
$<$10\% & \textit{\textbf{PaleSkin}} & 0.12 & 0.20 & \textbf{0.37} \\ 
& \textit{\textbf{Mustache}} & 0.39 & 0.30 & \textbf{0.46} \\ 
& \textit{\textbf{Bald}} & 0.88 & 1.20 & \textbf{2.18} \\ 
& \textit{\textbf{WearingNecktie}} & 0.96 & 0.80 & \textbf{1.23} \\ 
& \textit{\textbf{PointyNose}} & 1.42 & 1.80 & \textbf{2.95} \\ 
& \textit{\textbf{GrayHair}} & 2.43 & 1.90 & \textbf{3.12} \\ 
& \textit{\textbf{RecedingHairline}} & 3.32 & 2.40 & \textbf{3.92} \\ 
& \textit{\textbf{BigLips}} & 4.66 & 5.90 & \textbf{7.26} \\ 
& \textit{\textbf{BlondHair}} & 5.04 & 5.20 & \textbf{5.72} \\ 
& \textit{\textbf{Eyeglasses}} & 6.10 & 5.70 & \textbf{6.51} \\ 
& \textit{\textbf{WearingHat}} & 8.07 & 7.80 & \textbf{11.03} \\ 
\hline
$>$10\% & \textit{\textbf{BrownHair}} & 19.76 & \textbf{22.92} & 13.78 \\ \hline
\end{tabular}
\caption{Percentage of LFWA attributes predicted by FaceXformer for Section 4.1. Sorted in descending order of FFHQ(\%). The entire table is in Table~\ref{tab:11}.}
\label{tab:2}
\end{table}

\subsubsection{Generated Rare Facial Attributes}
Our method successfully increases the percentages of rare attributes as shown in Tables~\ref{tab:1} and~\ref{tab:2}.
To identify rare facial attributes in FFHQ, we employ FaceXFormer \cite{narayan2024facexformer}, which provides multiple face-related features including age, gender, race, head pose, and the attributes from Deep Learning Face Attributes in the Wild (LFWA) dataset \cite{liu2015faceattributes}. Real data attributes with lower percentages include extreme ages (very young or old), male gender, non-white races, non-frontal head poses\footnote{The head pose is predicted in the form of $(\theta_1, \theta_2, \theta_3)=$(pitch, yaw, roll). We define \textit{Front} as the head pose with $-15^{\circ}<\theta_1,\theta_2, \theta_3<15^{\circ}$.}, non-natural skin colors, hairless or no hair, hair colors other than brown, and accessories.

\begin{figure}
    \centering
    \includegraphics[width=0.99\linewidth]{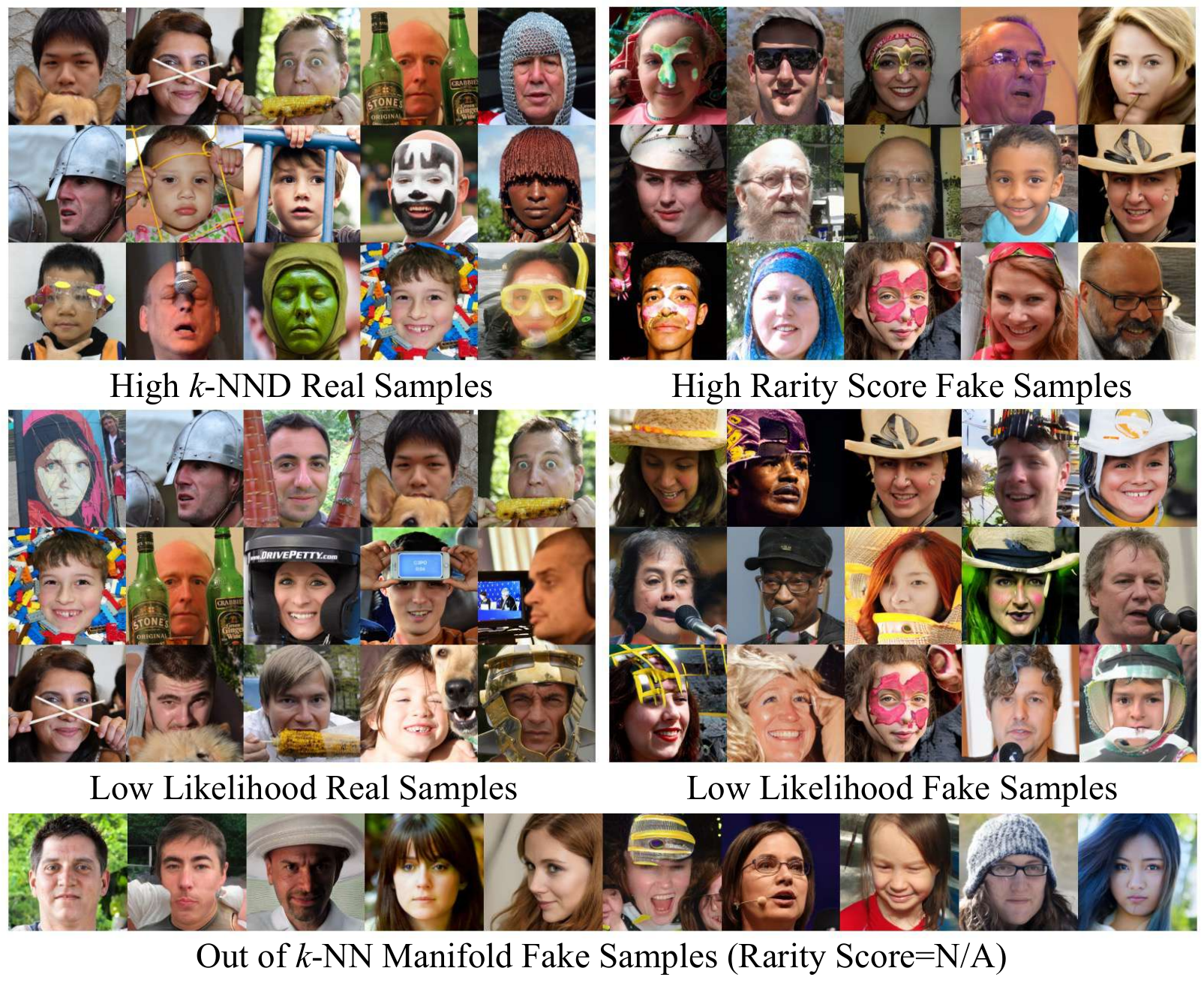}
    \caption{Examples of high- and low-density real and fake samples for Section 4.1.}
    \label{fig:3}
\end{figure}

Additionally, other rare attributes can be identified qualitatively. We select and visualize the top- and bottom-ranked samples based on $k$-NN-based and likelihood-based density estimates in Fig.~\ref{fig:3}. For real samples, $k$-NND \cite{loftsgaarden1965nonparametric} is employed, while the rarity score \cite{han2023rarity} is used for fake samples. Likelihoods are estimated using the NF model, excluding samples outside the real $k$-NN manifold. Rare samples exhibit characteristics such as objects obscuring faces, face painting, various hats, colorful eyeglasses, and artifacts. Notably, samples with undefined (N/A) rarity scores may include high-fidelity, artifact-free images, which arise from underestimated regions in the $k$-NN manifold.

\subsubsection{Qualitative Results}
\begin{figure}[t]
    \centering
    \includegraphics[width=0.99\linewidth]{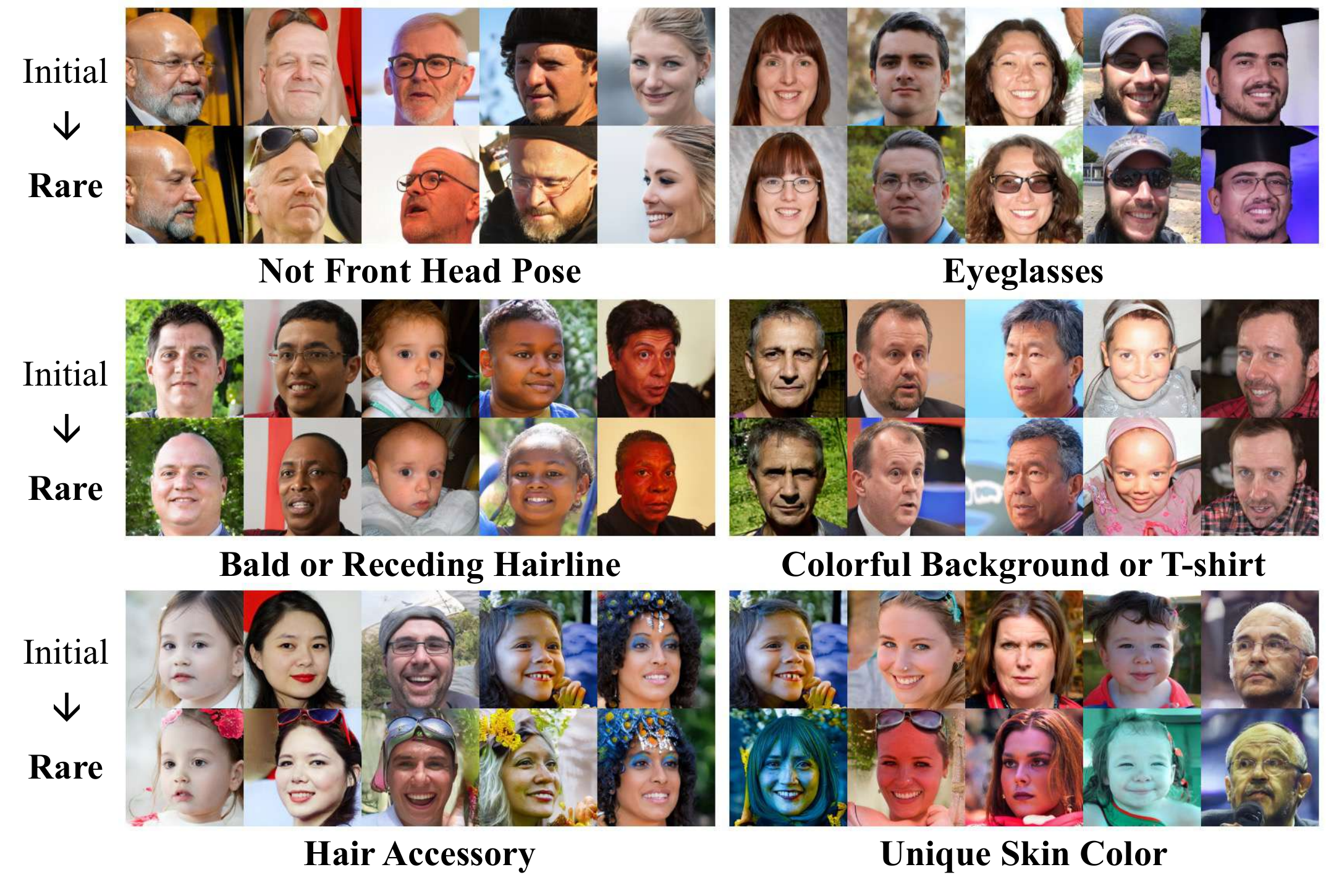}
    \caption{Examples of rare samples generated by our method for Section 4.1.}
    \label{fig:4}
\end{figure}

\begin{figure}[t]
    \centering
    \includegraphics[width=0.99\linewidth]{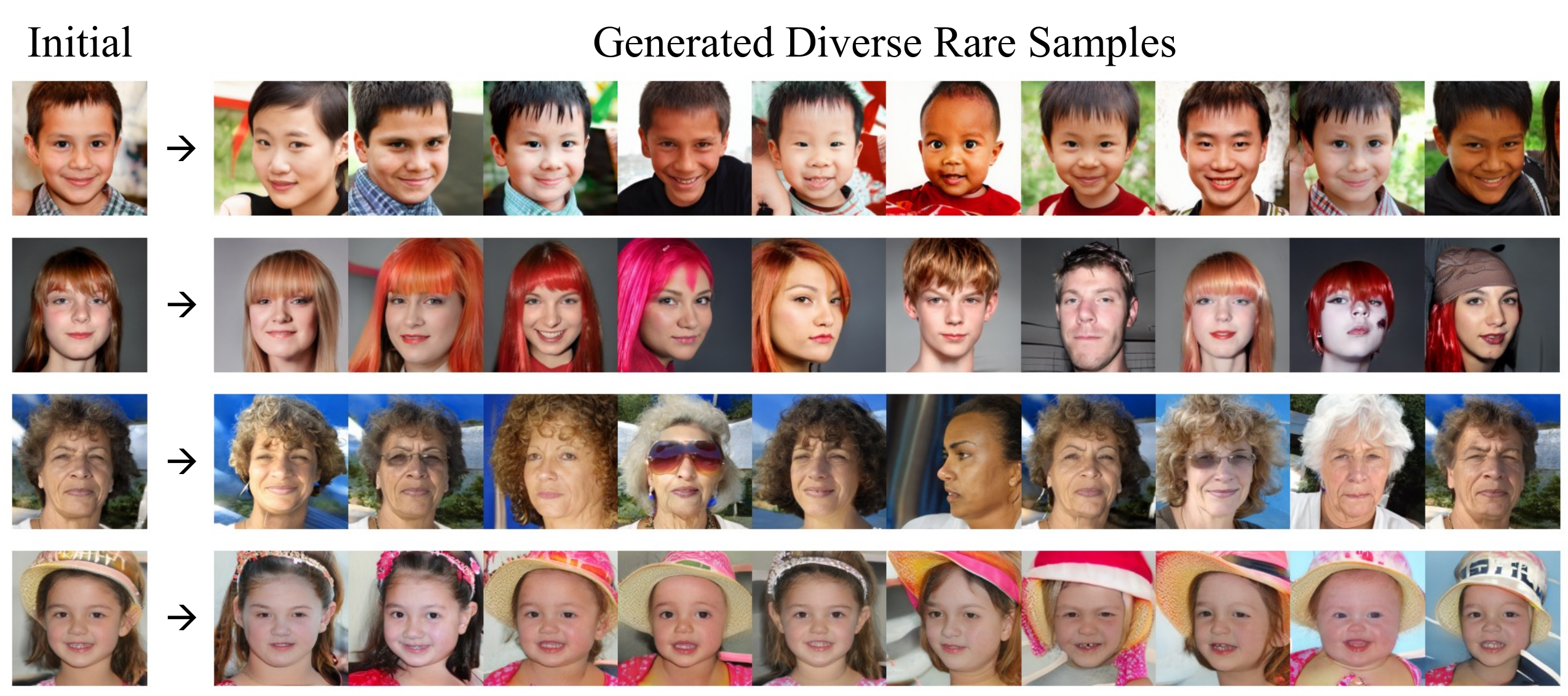}
    \caption{Examples of diverse rare samples generated by our method for Section 4.1.}
    \label{fig:5}
\end{figure}

As shown in Fig.~\ref{fig:1}~(Right) and~\ref{fig:4}, our method generates samples with rare attributes, including hats, hair colors other than natural brown, very young or old age, non-white races such as Black, Indian, and Asian, non-frontal head poses, eyeglasses, bald or receding hairline, colorful backgrounds or T-shirts, hair accessories, and unique skin colors. Moreover, the rare samples generated by our method show diversity, as shown in Fig.~\ref{fig:1}~(Left) and~\ref{fig:5}. Starting from initial vectors with small noise variations, we generate diverse rare images that retain perceptual similarity to the reference. Additional examples are provided in Appendix D.

\subsection{Animal Face and Artwork Generation with StyleGAN2-ADA}
\subsubsection{Quantitative \& Qualitative Results}
As a baseline, 5,000 synthetic samples are generated using latent vectors from StyleGAN2-ADA with a truncation parameter of $\psi=1.0$. With our method, we generate five rare samples for each of 1,000 initial latent vectors from the baseline, using parameters $\lambda_1=200.0, \lambda_2=0.02, \sigma=0.01$ and $k'=100$.

\begin{figure}
    \centering
    \includegraphics[width=0.99\linewidth]{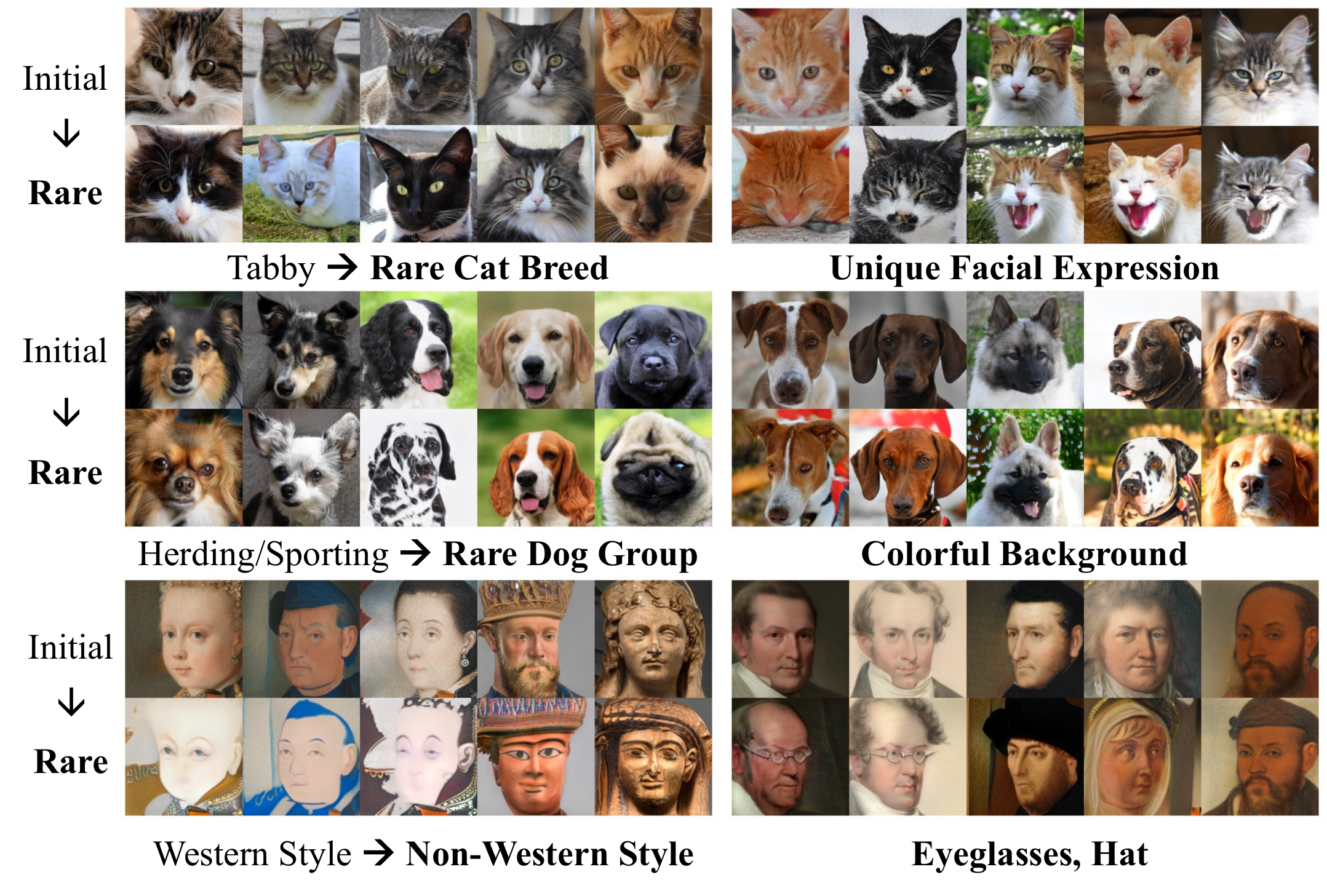}
    \caption{Examples of rare samples generated by our method for Section 4.2.}
    \label{fig:7}
\end{figure}

We also evaluate the results using various metrics, as presented in Table~\ref{tab:6}. Given that the AFHQ and MetFaces datasets are relatively small, we use the KID score \cite{binkowski2018demystifying} instead of the FID score, as the KID score is inherently unbiased \cite{karras2020training}. Our method effectively enhances rarity and diversity across all three datasets compared to the baseline.
In Fig.~\ref{fig:7}, we visualize the examples generated by our method. More examples are in Fig.~\ref{fig:14} and~\ref{fig:15}.

\setlength{\tabcolsep}{0.6mm}
\begin{table}[t]
\centering
\begin{tabular}{ll|rrrrr}
\hline
\textbf{Data} & \textbf{Model} & \textbf{\begin{tabular}[c]{@{}l@{}}RS $\uparrow$\end{tabular}} & \textbf{Prec.} $\uparrow$ & \textbf{Rec.} $\uparrow$ & \textbf{LPIPS} $\uparrow$ & 
\textbf{\begin{tabular}[c]{@{}l@{}}KID $\downarrow$\\\small $\times 10^3$\end{tabular}}\\ 
\hline
\multirow{2}{*}{\textbf{AFHQ Cat}} & \textbf{Ref.}  & 17.47 & \textbf{0.76} & 0.49 & 0.73 & \textbf{0.46}  \\ 
& \textbf{Ours}  & \textbf{20.71} & \textbf{0.76} & \textbf{0.68} & \textbf{0.78 } & 2.03 \\ 
\hline
\multirow{2}{*}{\textbf{AFHQ Dog}} & \textbf{Ref.} & 21.19 & 0.77 & 0.56& 0.75 & \textbf{1.10}    \\ 
& \textbf{Ours} & \textbf{27.32} & \textbf{0.85} & \textbf{0.76} & \textbf{0.80} & 2.73  \\ 
\hline
\multirow{2}{*}{\textbf{MetFaces}} & \textbf{Ref.} & 16.90 & 0.80 & 0.44 & 0.74 & \textbf{0.97}   \\ 
& \textbf{Ours} & \textbf{21.25} & \textbf{0.86} & \textbf{0.62} & \textbf{0.78} & 2.15 \\ 
\hline
\end{tabular}
\caption{Quantitative evaluation for Section 4.2.}
\label{tab:6}
\end{table}

\begin{figure}
    \centering
    \includegraphics[width=0.99\linewidth]{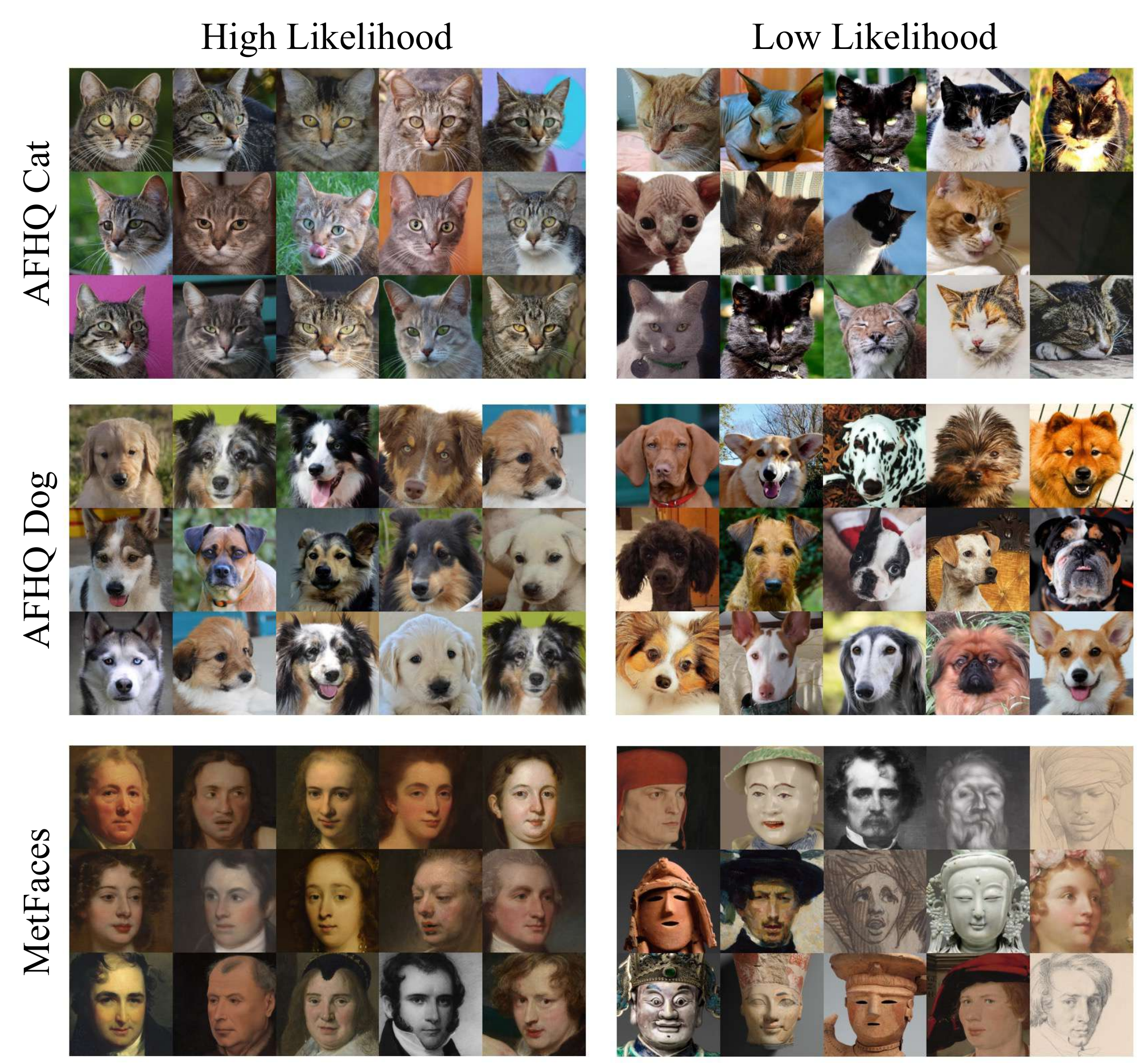}
    \caption{Examples of high- and low-likelihood real samples from the AFHQ Cat, Dog, and MetFaces dataset.}
    \label{fig:6}
\end{figure}

\paragraph{Generated Rare Attributes of Animal Face and Artwork}
To identify the rare cat and dog breeds in AFHQ datasets, we employed Model Soups \cite{wortsman2022model} for zero-shot classification of cat and dog classes in the ImageNet dataset \cite{deng2009imagenet}, which includes five cat classes and 120 dog classes. In the AFHQ Dog dataset, each dog class represents less than 5\% of the total, so we grouped the dogs into broader categories based on appearance: Toy, Hound, Scent Hound, Terrier, Sporting, Non-Sporting, Herding, and Working. Further details are provided in Appendix E. The classified result is shown in Table~\ref{tab:4}, and our method successfully increases the percentages of minor classes.

We also apply the FaceXFormer and observe similar rare attributes in FFHQ, as shown in Table~\ref{tab:5}.

To further identify rare attributes within the datasets, we visualize the high- and low-likelihood samples in Fig.~\ref{fig:6}. For the AFHQ-cat dataset, high-likelihood samples predominantly consist of brown-colored Tabby cats, whereas low-likelihood samples encompass a broader range of classes. In the AFHQ-dog dataset, high-likelihood samples are primarily drawn from the Herding and Sporting groups, including breeds such as Shetland Sheepdogs, Collies, and Retrievers. In contrast, low-likelihood samples span a variety of groups and exhibit greater diversity in head poses, backgrounds, and facial expressions. In the Metfaces dataset, high-likelihood samples predominantly include European-style oil paintings, while low-likelihood samples include statues, drawings, and other styles of paintings. As shown in Fig.~\ref{fig:7}, such rare attributes can be also observed in the results of our method.

\setlength{\tabcolsep}{2mm}
\begin{table}
\centering
\begin{tabular}{ll|rrr}
\hline
\multicolumn{2}{l|}{\textbf{ImageNet Attribute}} & \textbf{Real} & \textbf{Reference} & \textbf{Ours(\%)} \\ 
\hline
\textbf{Cat} & \textbf{Tabby} & 56.76 & \textbf{58.00} & 55.10   \\
& \textbf{Others} & 43.23 & 42.00 & \textbf{44.90}    \\ 
\hline
\textbf{Dog} & \textbf{Herding} & 21.48 & \textbf{21.70} & 17.68   \\
& \textbf{Sporting} & 18.50 & \textbf{17.30} & 15.24   \\
& \textbf{Working} & 15.44 & 14.60 & \textbf{16.84}  \\
& \textbf{Toy} & 14.93 & 17.30 & \textbf{19.44}  \\
& \textbf{Terrier} & 9.11 & \textbf{9.60} & 7.72    \\
& \textbf{Non-Sporting}& 7.02 & 5.90 & \textbf{7.38}   \\
& \textbf{Scent Hound} & 6.92 & 8.10 & \textbf{8.92}   \\
& \textbf{Hound} & 3.05 & 2.60 & \textbf{3.76}   \\ 
\hline
\end{tabular}
\caption{Percentage of the cat-related breeds and dog-related groups in ImageNet classes predicted by Model Soups. Sorted in descending order of Real(\%). The entire table is in Table~\ref{tab:12}.}
\label{tab:4}
\end{table}

\setlength{\tabcolsep}{1.4mm}
\begin{table}
\centering
\begin{tabular}{ll|rrr}
\hline
\multicolumn{2}{l|}{\textbf{Model-Based Attribute}} & \textbf{MetF.} & \textbf{Reference} & \textbf{Ours(\%)} \\ 
\hline
\multirow{3}{*}{\textbf{Age}}  
& \textit{\textbf{0-9}}      & 5.19  & 3.23 & \textbf{5.21}     \\ 
& \textit{\textbf{10-69}}    & 94.20 & \textbf{95.96} & 93.55   \\ 
& \textit{\textbf{Over70}}   & 0.60  & 0.80  & \textbf{1.23}     \\ 
\hline
\multirow{2}{*}{\textbf{Gender}} 
& \textit{\textbf{Male}}     & 42.84 & 41.71 & \textbf{49.62}   \\ 
& \textit{\textbf{Female}}   & 57.15 & \textbf{58.28} & 50.37   \\ 
\hline
\multirow{5}{*}{\textbf{Race}} 
& \textit{\textbf{White}}    & 73.64 & \textbf{77.97} & 77.96   \\ 
& \textit{\textbf{Indian}}   & 9.86  & \textbf{9.19}  & 9.03    \\ 
& \textit{\textbf{Black}}    & 3.38  & 2.02 & \textbf{2.33}     \\ 
& \textit{\textbf{Asian}}    & 2.86  & 4.04 & \textbf{5.25}     \\ 
& \textit{\textbf{Others}}   & 10.24 & \textbf{6.76} & 5.40     \\ 
\hline
\multirow{2}{*}{\textbf{HeadPose}} 
& \textit{\textbf{Front}}    & 42.54 & \textbf{46.06} & 32.86   \\ 
& \textit{\textbf{NotFront}} & 57.45 & 53.93 & \textbf{67.14}   \\ 
\hline
\multirow{2}{*}{\textbf{LFWA}} 
& \textit{\textbf{Eyeglasses}} & 0.07 & 0.10 & \textbf{0.33}    \\ 
& \textit{\textbf{WearingHat}} & 12.34 & 9.89 & \textbf{12.18}  \\ 
\hline
\end{tabular}
\caption{Percentage of age, gender, race, head pose, \textit{Eyeglasses}, and \textit{WearingHat} attributes predicted by FaceXFormer. MetF. refers to the MetFaces dataset. The entire table is in Table~\ref{tab:17}.}
\label{tab:5}
\end{table}

\subsection{Ablation Study on the Objective Function}
Our objective function includes three components: rarity, similarity, and diversity terms. To evaluate their effectiveness, we conduct an ablation study using the FFHQ dataset and StyleGAN2, keeping the density estimator and parameters consistent. We optimize ten samples for each of the 100 initial latent vectors across different objective combinations.

First, we assess results using only the $\mathcal{L}_{rare}$. Adding the $\mathcal{L}_{sim}$ ensures that samples stay within a similarity boundary to the reference, potentially finding rarer and more diverse samples inside the boundary. Finally, incorporating the $\mathcal{L}_{div}$ completes the full objective. The results in Table~\ref{tab:7} demonstrate the effectiveness of each term.

\setlength{\tabcolsep}{3.5mm}
\begin{table}
\centering
\begin{tabular}{l|rr}
\hline
\textbf{Objective}  & \textbf{RS} & \textbf{LPIPS} \\ \hline
$\mathcal{L}_{rare}$        & 18.99 & 0.752 \\
$\mathcal{L}_{rare}+\lambda_1\mathcal{L}_{sim}$ & 21.11 & 0.766 \\
$\mathcal{L}_{rare}+\lambda_1\mathcal{L}_{sim}+\lambda_2\mathcal{L}_{div}$ & \textbf{21.28} & \textbf{0.768} \\ \hline
\end{tabular}
\caption{Rarity scores (RS) and LPIPS scores for the ablation study on the objective function.}
\label{tab:7}
\end{table}

\subsection{Relationship with Rarity Score}
\begin{figure}
    \centering
    \begin{minipage}{0.5\linewidth} 
        \centering
        \includegraphics[width=0.99\linewidth]{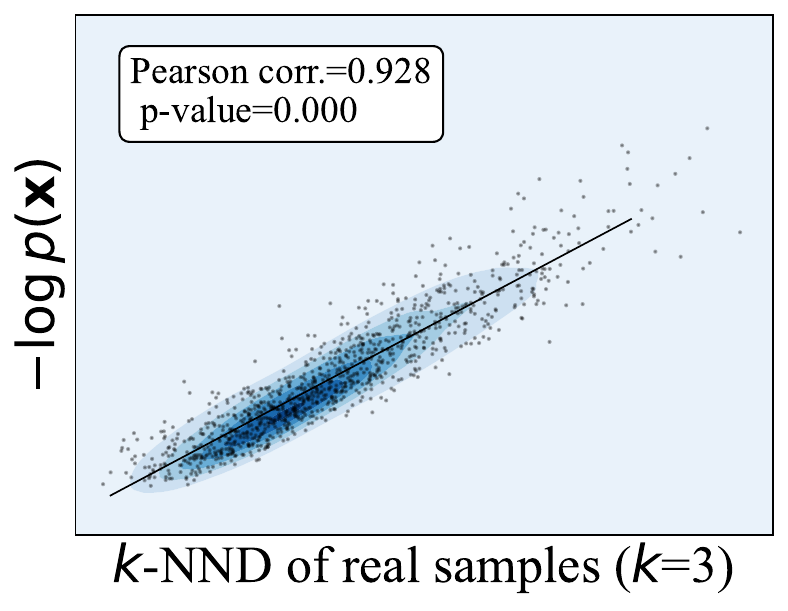}
    \end{minipage}
    \begin{minipage}{0.46\linewidth} 
        \centering
        \includegraphics[width=0.99\linewidth]{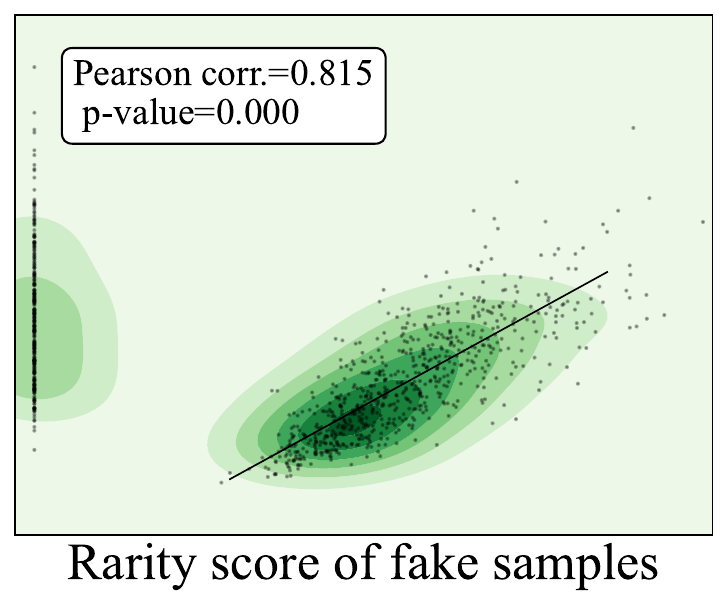}
    \end{minipage}
    \caption{Correlation plot for the $k$-NND / rarity score and negative log-likelihood estimated by the normalizing flow.}
    \label{fig:8}
\end{figure}

We use the rarity score \cite{han2023rarity} to measure sample rarity and demonstrate that our method improves rarity compared to other sampling methods. Although directly optimizing the rarity score is challenging, our likelihood-based objective effectively guides samples to locally low-density regions. To compare $k$-NN-based density measures ($k$-NND for real samples and rarity score for fake samples) with NF-estimated density measures, we visualize the scatter plot and compute the Pearson correlation coefficient as represented in Fig.~\ref{fig:8}. We observe a high Pearson correlation coefficient of 0.928 for real samples and 0.815 for fake samples, with p-values $< 10^{-8}$, excluding samples with undefined rarity scores.

\begin{figure}
    \centering
    \includegraphics[width=0.93\linewidth]{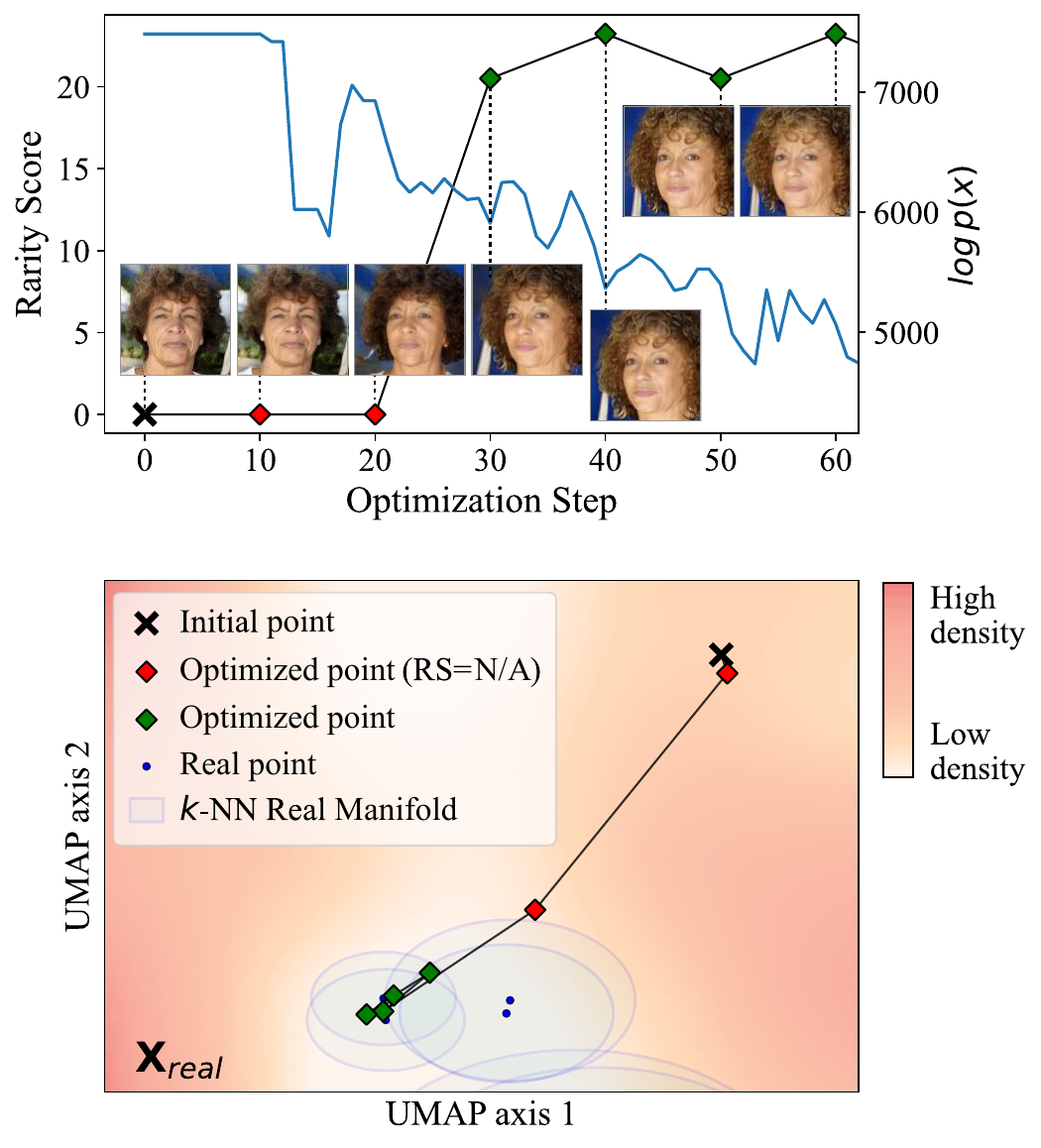}
    \caption{Example of the optimization path with a real $k$-NN manifold and a heatmap of likelihoods estimated by the normalizing flow. Notably, the local $k$-NN manifold includes only the three nearest real data points ($k=3$) for each point, rather than the entire manifold.}
    \label{fig:9}
\end{figure}

Although the NF estimates likelihood across the feature space, the rarity score is undefined outside the real $k$-NN manifold. This allows for out-of-manifold samples with sufficient quality, as shown at the bottom of Fig.~\ref{fig:3}. In Fig.~\ref{fig:9}, we visualize an optimization example that starts with an undefined rarity score but eventually gains and increases the rarity score. The sample, initially in an underestimated $k$-NN region, becomes rare by moving to a low-likelihood region, altering the reference image to achieve curlier blonde hair and a non-frontal head pose. We plot the NF-estimated density using RBF kernel interpolation and UMAP \cite{mcinnes2018umap} dimensionality reduction on the real feature space and its inverse transformation function. Further details are provided in Appendix F.

\section{Conclusion}
We proposed a novel algorithm that generates diverse rare samples using multi-start gradient-based optimization, avoiding low-quality samples. Users can control rarity, diversity, and similarity to the reference through a multi-objective approach. Our method successfully increased the prevalence of rare attributes in various image generation domains. We also provide an experimental comparison between $k$-NN-based and normalizing flow-based density estimation methods. We hope this work contributes to advancing creativity in deep generative models. 
However, there are some limitations that could be improved. The results rely on the GAN's capabilities and require an additional density estimator. Exploring other generative models might improve outcomes and eliminate the need for extra training. Additionally, our method alters multiple attributes simultaneously; integrating it with other image manipulation techniques could allow for more controlled manipulation.

\section*{Acknowledgments}
This work was partly supported by KAIST-NAVER Hypercreative AI Center, and from the Korean Institute of Information \& Communications Technology Planning \& Evaluation and the Korean Ministry of Science and ICT under grant agreement No. RS-2019-II190075 (Artificial Intelligence Graduate School Program(KAIST)), No. RS-2022-II220984 (Development of Artificial Intelligence Technology for Personalized Plug-and-Play Explanation and Verification of Explanation), and No.RS-2022-II220184 (Development and Study of AI Technologies to Inexpensively Conform to Evolving Policy on Ethics).

\small\bibliography{aaai25}
\normalsize\clearpage
\appendix
\section{$k$-NN-based Evaluation Metrics}
$k$-NN-based manifold estimation is employed in various metrics to assess both fidelity and diversity of synthetic samples \cite{kynkaanniemi2019improved, naeem2020reliable}.     
For real samples $\mathbf{I}_r\sim P_r$ and fake samples $\mathbf{I}_g \sim P_g$, they are embedded in the feature space with the pretrained DNNs such as VGG16 ~\citep{simonyanZ14a} or the CLIP image encoder ~\citep{radford2021learning} to get sets of feature vectors $\mathbf{X_r}$ and $\mathbf{X_g}$, respectively. The real and fake manifolds are estimated by the given sample sets as follows. 
\begin{equation}
\begin{aligned}
\Phi_\textbf{X} &= \bigcup_{\mathbf{x}_i \in \mathbf{X}} B_k(\mathbf{x}_i,\mathbf{X})\\ 
B_k(\mathbf{x}_i, \mathbf{X}) &=\{\mathbf{x}| d(\mathbf{x}_i,\mathbf{x}) \le k\text{-NND}(\mathbf{x}_i,\mathbf{X})\}
\end{aligned}
\end{equation}
Here, $k\text{-NND}(\mathbf{x}_i,\mathbf{X})$ represents the distance between $\mathbf{x}_i$ and its $k$-th nearest neighbor in $\mathbf{X}$. $B_k(\mathbf{x}_i,\mathbf{X})$ is the $k$-NN ball (hyper-sphere) with the radius of $k\text{-NND}(\mathbf{x}_i,\mathbf{X})$ centered at $\mathbf{x}_i$ defined as a set of all $\mathbf{x}$ whose distance to $\mathbf{x}_i$ is smaller than or equal to $k\text{-NND}(\mathbf{x}_i,\mathbf{X})$. For simplicity, we use $\Psi_{real}=\Psi_{X_r}$ through the paper.

We utilize three $k$-NN-based evaluation metrics in the quantitative analysis, precision, recall, and rarity score. Precision \cite{kynkaanniemi2019improved} measures the proportion of fake samples within the real manifold, indicating how realistic the fake samples are. Recall \cite{kynkaanniemi2019improved} measures the proportion of real samples within the fake manifold, assessing how well the generative model captures the modes of the real data distribution. The rarity score \cite{han2023rarity} measures the uniqueness of individual samples, with the following formulation.  
\begin{equation}\label{eq:score}
    \text{rarity}(\mathbf{x}_g,\mathbf{X_r}) = \min_{r,\,\, s.t. \mathbf{x}_g\in B_k(\mathbf{x}_r,\mathbf{X_r}) }k\text{-NND}(\mathbf{x}_r,\mathbf{X_r}).
\end{equation}

\section{Implementation Details}
\subsection{Computational Resources}
For all experiments including model training and inference, and optimization, we utilized a single NVIDIA RTX A6000 GPU with PyTorch version of 2.1.0+cu121.
\subsection{Normalizing Flow Architecture}
We use the Glow architecture proposed by \citet{kingma2018glow} for our density estimation model, adapting it from its original design for RGB images to work in a feature space with a dimension of $\mathbb{R}^{4096}$, representing the second-to-last latent space of the VGG16-fc2 model \cite{simonyan2014very}. While retaining the original Glow structure—stacked blocks of sequential flows with Actnorm, invertible $1\times1$ convolution, affine coupling, and split layers—we modify Actnorm layers for grouped channel-wise operations, since our feature vector lacks a patch-like structure. To achieve this, we introduce a $1\times1$ convolution layer for general permutation before each Actnorm layer, followed by dividing the dimensions into a user-defined number of groups. This modification makes the model lighter and more efficient. The finalized architecture is represented in Table~\ref{tab:8}.

We train the NF model with 70,000 samples for FFHQ, 5,653 for AFHQ Cat, 5,239 for AFHQ Dog, and 1,336 for MetFaces, splitting FFHQ and MetFaces images 7:3 for training and validation, and using the provided original splits for the AFHQ datasets. We use a batch size of 32, scale data to $[0,1]$ by a min-max scaler, and apply the Adam optimizer \cite{diederik2014adam} with a learning rate of $1\times10^{-4}$ and the StepLR scheduler with a step size of 500, gamma of 0.1. The number of flows, blocks, and groups for the modified Actnorm are 32, 4, and 4, respectively. The best checkpoint was obtained at 3,000, 2,000, and 1,500 iterations for FFHQ, AFHQ, and MetFaces. Training takes less than 30 minutes on a single GPU.

\setlength{\tabcolsep}{0.5mm}
\begin{table}
    \centering
    \begin{tabular}{c|c|c|c|c}
    \hline
    \textbf{Block} & \multicolumn{2}{c|}{\textbf{Layer}} & \textbf{Input dim} & \textbf{Output collect} 
    \\ \cline{1-5}
    \multirow{6}{*}{\textbf{0}} 
    & \multicolumn{2}{c|}{$1 \times 1$ conv} & 1 $\times$ 4096 & - 
    \\ \cline{2-5}
    & \multicolumn{2}{c|}{grouping} & 4 $\times$ 1024 & - 
    \\ \cline{2-5}
    & \multirow{3}{*}{\begin{tabular}[c]{@{}c@{}}flow \\($\times$ 32)\end{tabular}} & actnorm & 4 $\times$ 1024 & - 
    \\
    & & $1 \times 1$ conv & 4 $\times$ 1024 & - 
    \\
    & & affine coupling & 4 $\times$ 1024 & - 
    \\ \cline{2-5}
    & \multicolumn{2}{c|}{split} & 2 $\times$ 1024 & yes 
    \\ \cline{1-5}
    
    \multirow{6}{*}{\textbf{1}} 
    & \multicolumn{2}{c|}{$1 \times 1$ conv} & 2 $\times$ 1024 & - 
    \\ \cline{2-5}
    & \multicolumn{2}{c|}{grouping} & 8 $\times$ 256 & - 
    \\ \cline{2-5}
    & \multirow{3}{*}{\begin{tabular}[c]{@{}c@{}}flow \\($\times$ 32)\end{tabular}} & actnorm & 8 $\times$ 256 & - 
    \\
    & & $1 \times 1$ conv & 8 $\times$ 256 & - 
    \\
    & & affine coupling & 8 $\times$ 256 & - 
    \\ \cline{2-5}
    & \multicolumn{2}{c|}{split} & 4 $\times$ 256 & yes 
    \\ \cline{1-5}
    
    \multirow{6}{*}{\textbf{2}} 
    & \multicolumn{2}{c|}{$1 \times 1$ conv} & 4 $\times$ 256 & - 
    \\ \cline{2-5}
    & \multicolumn{2}{c|}{grouping} & 16 $\times$ 64 & - 
    \\ \cline{2-5}
    & \multirow{3}{*}{\begin{tabular}[c]{@{}c@{}}flow \\($\times$ 32)\end{tabular}} & actnorm & 16 $\times$ 64 & - 
    \\
    & & $1 \times 1$ conv & 16 $\times$ 64 & - 
    \\
    & & affine coupling & 16 $\times$ 64 & - 
    \\ \cline{2-5}
    & \multicolumn{2}{c|}{split} & 8 $\times$ 64 & yes 
    \\ \cline{1-5}
    
    \multirow{6}{*}{\textbf{3}} 
    & \multicolumn{2}{c|}{$1 \times 1$ conv} & 8 $\times$ 64 & - 
    \\ \cline{2-5}
    & \multicolumn{2}{c|}{grouping} & 32 $\times$ 16 & - 
    \\ \cline{2-5}
    & \multirow{3}{*}{\begin{tabular}[c]{@{}c@{}}flow \\($\times$ 32)\end{tabular}} & actnorm & 32 $\times$ 16 & - 
    \\
    & & $1 \times 1$ conv & 32 $\times$ 16 & - 
    \\
    & & affine coupling & 32 $\times$ 16 & yes 
    \\ 
    \hline
    \end{tabular}
    \caption{Normalizing flow model architecture.}
    \label{tab:8}
\end{table}

\subsection{Settings for Optimization}
For the diverse rare sample optimization, we use a maximum number of 200 epochs. We use the StepLR scheduler from the PyTorch package for learning rates, with the gamma of 0.9. The step sizes are set to 50 for the FFHQ-StyleGAN2 experiments and 100 for the AFHQ-StyleGAN2-ADA and MetFaces-StyleGAN2-ADA experiments. For other parameters, the default settings are used.

In the FFHQ-StyleGAN2 experiments, the average time of the optimization for the one initial latent vector with $N=10$ is less than 8 minutes with a single GPU.

\setlength{\tabcolsep}{2.2mm}
\begin{table}[]
\begin{tabular}{ll|rrr}
\hline
\textbf{$\lambda_1$} & \textbf{$\lambda_2$} & \textbf{Rarity Score$\uparrow$} & \textbf{LPIPS$\uparrow$} &\textbf{OOM$\downarrow$(\%)} \\ \hline
\textbf{20}          & \textbf{0.002}       & 23.92                 & \textbf{0.769}  & 11.92        \\
\textbf{30}          & \textbf{0.002}       & 24.02                 & 0.768   & \textbf{11.40}      \\
\textbf{40}          & \textbf{0.002}       & \textbf{24.04}                 & 0.767     & 11.79    \\
\textbf{30}          & \textbf{0.001}       & 23.92                 & 0.766   & 12.10      \\
\textbf{30}          & \textbf{0.003}       & 23.87                      & 0.767    & 13.00          \\
\textbf{30}          & \textbf{0.005}       & 23.72                 & \textbf{0.769}    & 12.20     \\ \hline
\end{tabular}
\caption{Rarity score and LPIPS score from our method with varying $\lambda_1$ and $\lambda_2$, experimenting on FFHQ-StyleGAN2. OOM refers to out-of-manifold sample percentage.}
\label{tab:9}
\end{table}

\setlength{\tabcolsep}{1.6mm}
\begin{table}[]
\begin{tabular}{l|rrrr}
\hline
\textbf{$k'$} & \textbf{Rarity Score$\uparrow$} & \textbf{LPIPS$\uparrow$} & \textbf{LPIPS*$\downarrow$} & \textbf{OOM$\downarrow$(\%)} \\ \hline
\textbf{80}   & 23.78                 & 0.767    & \textbf{0.523}  & 13.60  \\
\textbf{100}  & 24.02                 & 0.768    & 0.532  & 11.40 \\
\textbf{200}  &  \textbf{24.23}                     &  \textbf{0.771}      &  0.546  &\textbf{10.70}  \\ \hline
\end{tabular}
\caption{Rarity score and LPIPS scores from our method with varying $k'$. LPIPS: Mean LPIPS score among the optimized samples. LPIPS*: Mean LPIPS score between reference and the optimized samples, experimenting on FFHQ-StyleGAN2. OOM refers to out-of-manifold sample percentage.}
\label{tab:10}
\end{table}

\section{Choice of Parameters}
\paragraph{Coefficients for Objective Function $\lambda_1$ and $\lambda_2$}
We can control the strength of similarity regularization and diversity by adjusting the coefficients $\lambda_1$ and $\lambda_2$, respectively. A larger $\lambda_1$ encourages samples to remain within the penalizing boundary, potentially resulting in rarer and more diverse samples. However, if $\lambda_1$ is set too large, it may restrict diversity. On the other hand, increasing $\lambda_2$ enhances diversity, but if $\lambda_2$ is too large, it may reduce rarity due to the trade-offs inherent in the multi-objective framework. 

Varying these coefficients in the FFHQ-StyleGAN2 experimental setting, we visualize the mean rarity score except for the undefined rarity cases and pairwise LPIPS score in Table~\ref{tab:9}. These scores are calculated on the 1,000 samples generated from our method with 100 initial latent vectors and $N=10$. The LPIPS scores are calculated on randomly selected 10,000 pairs. The rarity score increases as $\lambda_1$ increases while decreasing as $\lambda_2$ increases. The LPIPS score increases as $\lambda_2$ increases while decreasing as $\lambda_1$ increases. In the parameter ranges shown in Table~\ref{tab:9}, both the rarity scores and LPIPS scores are higher than those of the baseline, with only slight differences between the different parameters.

\begin{figure}
    \centering
    \includegraphics[width=0.99\linewidth]{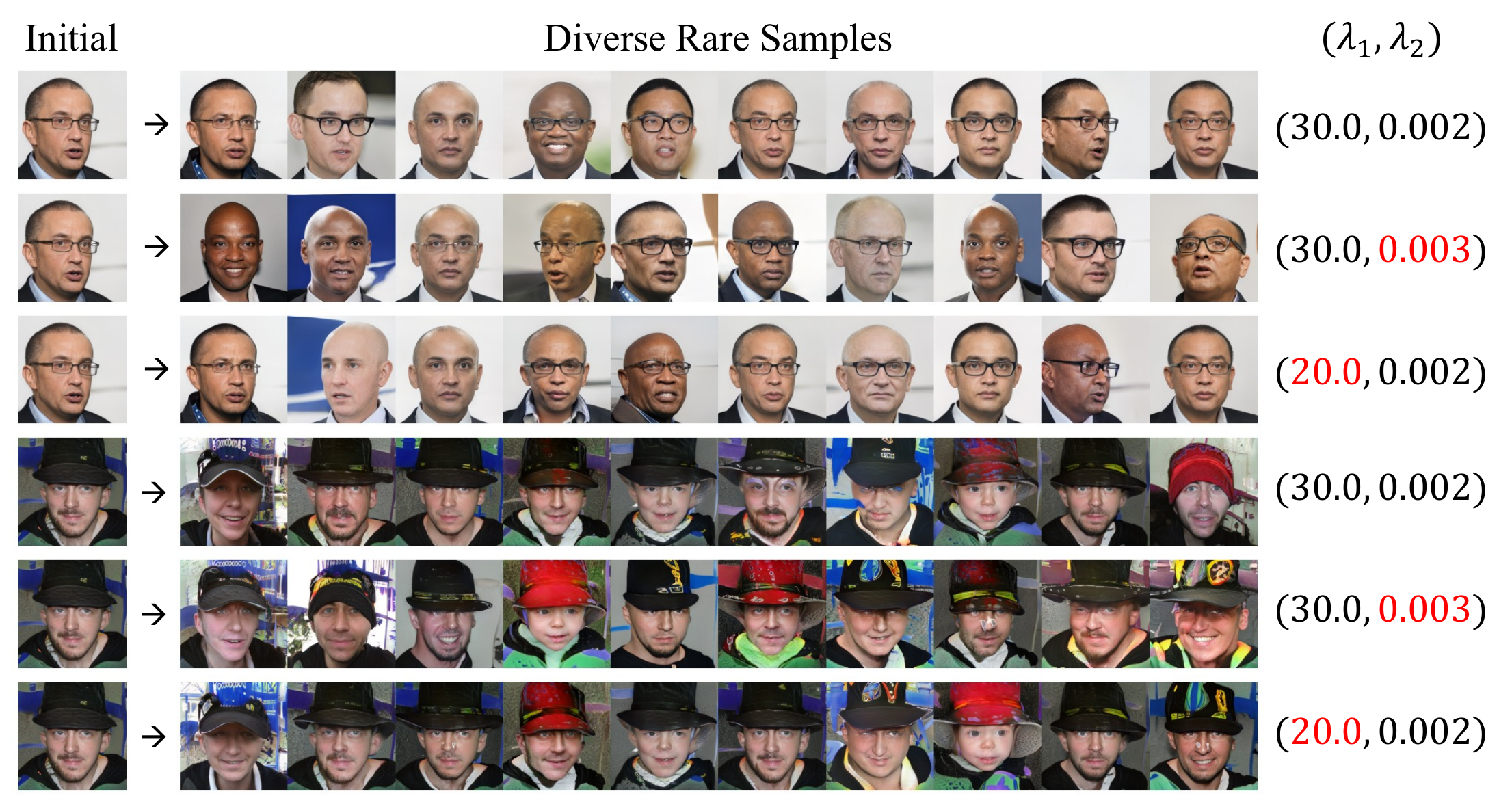}
    \caption{Examples of diverse rare samples generated by our method using FFHQ-StyleGAN2, varying the coefficients of the objective function, $\lambda_1$(similarity) and $\lambda_2$(diversity).}
    \label{fig:10}
\end{figure}

We unify the parameters for each dataset-model pair, however, there can be a better set of parameters for each reference. For instance, if we make $\lambda_2$ larger or $\lambda_1$ smaller for the samples with low diversity, the result can be more diverse. We provide the examples in Fig.~\ref{fig:10}. 

\paragraph{Parameter for the Penalizing Boundary $k'$}
The parameter $k'$ determines the radius of the penalizing boundary, which controls the similarity between the reference and optimized images. As $k'$ increases, the rarity and diversity of the optimized samples also increase, while the similarity between the reference and optimized images decreases, as shown in Table~\ref{tab:10}. We also use the LPIPS score to measure the similarity between the reference and the optimized image, with lower scores indicating greater similarity (denoted as LPIPS*).

\paragraph{Scale of Noise $\sigma$}
The first step in our diverse rare sample generation algorithm involves adding random noise to the initial latent vector to provide multi-starts for the optimization and promote diversity. This noise is sampled from the distribution $\mathcal{N}(\mathbf{0}, \sigma^2 I)$. While a larger $\sigma$ can yield more diverse results, setting $\sigma$ too high may produce out-of-distribution samples during the early stages of optimization. To balance diversity and realism, we select $\sigma$ values that result in fewer than 30\% of the initial perturbed latent vectors generating samples outside the real $k$-NN manifold. Specifically, 29.96\% for FFHQ-StyleGAN2, 22.40\% for AFHQ Cat-StyleGAN2-ADA, 19.00\% for AFHQ Dog-StyleGAN2-ADA, and 21.40\% for MetFaces-StyleGAN2-ADA. We allow some out-of-manifold samples in the initial stage since these may not be true out-of-distribution samples but rather appear as out-of-manifold due to the limitations of the $k$-NN-based manifold.

\setlength{\tabcolsep}{1.4mm}
\begin{table}[]
\begin{tabular}{l|rrr}
\hline
\textbf{Distance Metric} & \textbf{Rarity Score$\uparrow$} & \textbf{LPIPS$\uparrow$} &\textbf{OOM$\downarrow$(\%)} \\ \hline
\textbf{L2 (Original)}&24.02&0.768&11.40\\
\textbf{L1} &24.12&0.769&12.60\\
\textbf{Cosine Sim.}&24.00&0.769&13.80
\\ \hline
\end{tabular}
\caption{Rarity score and LPIPS score from our method with different distance metrics, experimenting on the FFHQ-StyleGAN2. OOM refers to out-of-manifold sample percentage.}
\label{tab:rebuttal_add}
\end{table}

\paragraph{Distance function $d$} We employ the Euclidean distance (L2 norm) in the feature space as the distance function for both similarity and diversity constraints, as described in Section 3.1. Experimental results using alternative distance metrics are presented in Table~\ref{tab:rebuttal_add}. In the FFHQ-StyleGAN2 setting, 1,000 samples are generated using our method with 100 initial latent vectors and $N=10$. Compared to results using the L1 norm and cosine similarity, our method achieves a higher rarity score and LPIPS score while generating fewer out-of-manifold samples than standard sampling, regardless of the metric used.

\setlength{\tabcolsep}{3.2mm}
\begin{table*}[t]
\centering
\begin{tabular}{lrrrlrrr}
\hline
\textbf{LFWA Attribute}
 &
  \textbf{FFHQ} &
  \textbf{Reference} &
  \textbf{Ours} &
   \textbf{LFWA Attribute}
  &\textbf{FFHQ} &
  \textbf{Reference} &
  \textbf{Ours(\%)} \\ \hline
\textit{\textbf{Blurry}} &0.005 &0 &\textbf{0.08} & \textit{\textbf{DoubleChin}} &11.14 &\textbf{10.41} &9.66 \\
\textit{\textbf{WearingNecklace}} &0.04 &0 &0 & \textit{\textbf{Bangs}} &12.04 &\textbf{12.31} &12.29 \\
\textit{\textbf{RosyCheeks}} &0.10 &\textbf{0.30} &0.11 & \textit{\textbf{BushyEyebrows}} &12.91 &\textbf{13.52} &10.30 \\
\textit{\textbf{PaleSkin}} &0.12 &0.20 &\textbf{0.37} & \textit{\textbf{WavyHair}} &13.12 &\textbf{15.31} &10.50 \\
\textit{\textbf{Mustache}} &0.39 &0.30 &\textbf{0.46} & \textit{\textbf{NarrowEyes}} &13.66 &12.11 &\textbf{13.59} \\
\textit{\textbf{Bald}} &0.88 &1.20 &\textbf{2.18} & \textit{\textbf{HeavyMakeup}} &14.97 &\textbf{16.51} &13.31 \\
\textit{\textbf{WearingNecktie}} &0.96 &0.80 &\textbf{1.23} & \textit{\textbf{Chubby}} &15.36 &13.81 &\textbf{13.92} \\
\textit{\textbf{PointyNose}} &1.42 &1.80 &\textbf{2.95} & \textit{\textbf{OvalFace}} &18.87 &\textbf{19.71} &12.40 \\
\textit{\textbf{GrayHair}} &2.42 &1.90 &\textbf{3.12} & \textit{\textbf{StraightHair}} &19.19 &17.81 &\textbf{19.65} \\
\textit{\textbf{RecedingHairline}} &3.32 &2.40 &\textbf{3.92} & \textit{\textbf{BrownHair}} &19.76 &\textbf{22.92} &13.78 \\
\textit{\textbf{ArchedEyebrows}} &4.43 &\textbf{4.20} &3.98 & \textit{\textbf{WearingLipstick}} &31.76 &\textbf{30.23} &26.91 \\
\textit{\textbf{BigLips}} &4.66 &5.90 &\textbf{7.26} & \textit{\textbf{Attractive}} &32.35 &\textbf{33.23} &29.64 \\
\textit{\textbf{BlondHair}} &5.04 &5.20 &\textbf{5.72} & \textit{\textbf{BagsUnderEyes}} &36.75 &\textbf{37.33} &36.04 \\
\textit{\textbf{Goatee}} &5.11 &\textbf{5.00} &4.52 & \textit{\textbf{BigNose}} &43.10 &43.94 &\textbf{45.31} \\
\textit{\textbf{Sideburns}} &5.63 &\textbf{5.40} &4.61 & \textit{\textbf{Male}} &47.51 &50.05 &\textbf{51.40} \\
\textit{\textbf{Eyeglasses}} &6.10 &5.70 &\textbf{6.51} & \textit{\textbf{HighCheekbones}} &55.55 &\textbf{52.85} &46.01 \\
\textit{\textbf{WearingEarrings}} &6.80 &\textbf{6.20} &4.85 & \textit{\textbf{Smiling}} &61.79 &\textbf{59.65} &54.56 \\
\textit{\textbf{WearingHat}} &8.07 &7.80 &\textbf{11.03} & \textit{\textbf{MouthSlightlyOpen}} &66.05 &\textbf{65.46} &63.62 \\
\textit{\textbf{BlackHair}} &8.09 &\textbf{8.90} &8.78 & \textit{\textbf{NoBeard}} &84.90 &85.68 &\textbf{86.45} \\
\textit{\textbf{5o'ClockShadow}} &10.90 &\textbf{10.91} &9.50 & \textit{\textbf{Young}} &85.21 &\textbf{85.18} &83.36 \\
\hline
\end{tabular}
\caption{Percentage of LFWA attributes predicted by FaceXFormer. 1,000 reference images are generated from FFHQ-StyleGAN2 with a truncation value of $\psi=1.0$, and the optimized images generated by our method are derived from the initial latent vectors of these 1,000 references with $N=10$. Sorted in ascending order of FFHQ(\%).}
\label{tab:11}
\end{table*}

\section{Additional Results for Section 4.1}
\subsection{Quantitative Results}
The full version of Table~\ref{tab:2} is provided in Table~\ref{tab:11}. Percentages are calculated only for face-detected cases, with undetected cases at 0.017\% for FFHQ, 0.100\% for references, and 0.529\% for optimized images.

Among the 19 rare attributes ($<$10\% in FFHQ), the percentages of 12 attributes increase with our method compared to the references. However, the percentages of six attributes—\textit{ArchedEyebrows, BlackHair, Goatee, RosyCheeks, Sideburns}, and \textit{WearingEarrings}—decrease due to being overshadowed by other rare attributes. Specifically, \textit{ArchedEyebrows, Goatee, Sideburns}, and \textit{WearingEarrings} often disappear when \textit{WearingHat} is present, while \textit{BlackHair} is replaced by other hair colors or \textit{Bald}. Additionally, FaceXFormer also frequently misses the \textit{RosyCheeks} attribute.

Attributes with high percentages ($>$40\%) in the dataset, such as \textit{BigNose} and \textit{NoBeard}, also increase with our method, which can be seemed weird. This result comes from the dependency on the likelihood estimated by the utilized NF model. To be specific, the rise in \textit{BigNose} may be due to its higher percentage among low-likelihood samples—49.39\% in the bottom 10\% versus 43.18\% in the entire sample set. \textit{NoBeard} increases as \textit{Goatee} and \textit{Sideburns} decrease.

\subsection{Qualitative Results}
\begin{figure}
    \centering
    \includegraphics[width=0.99\linewidth]{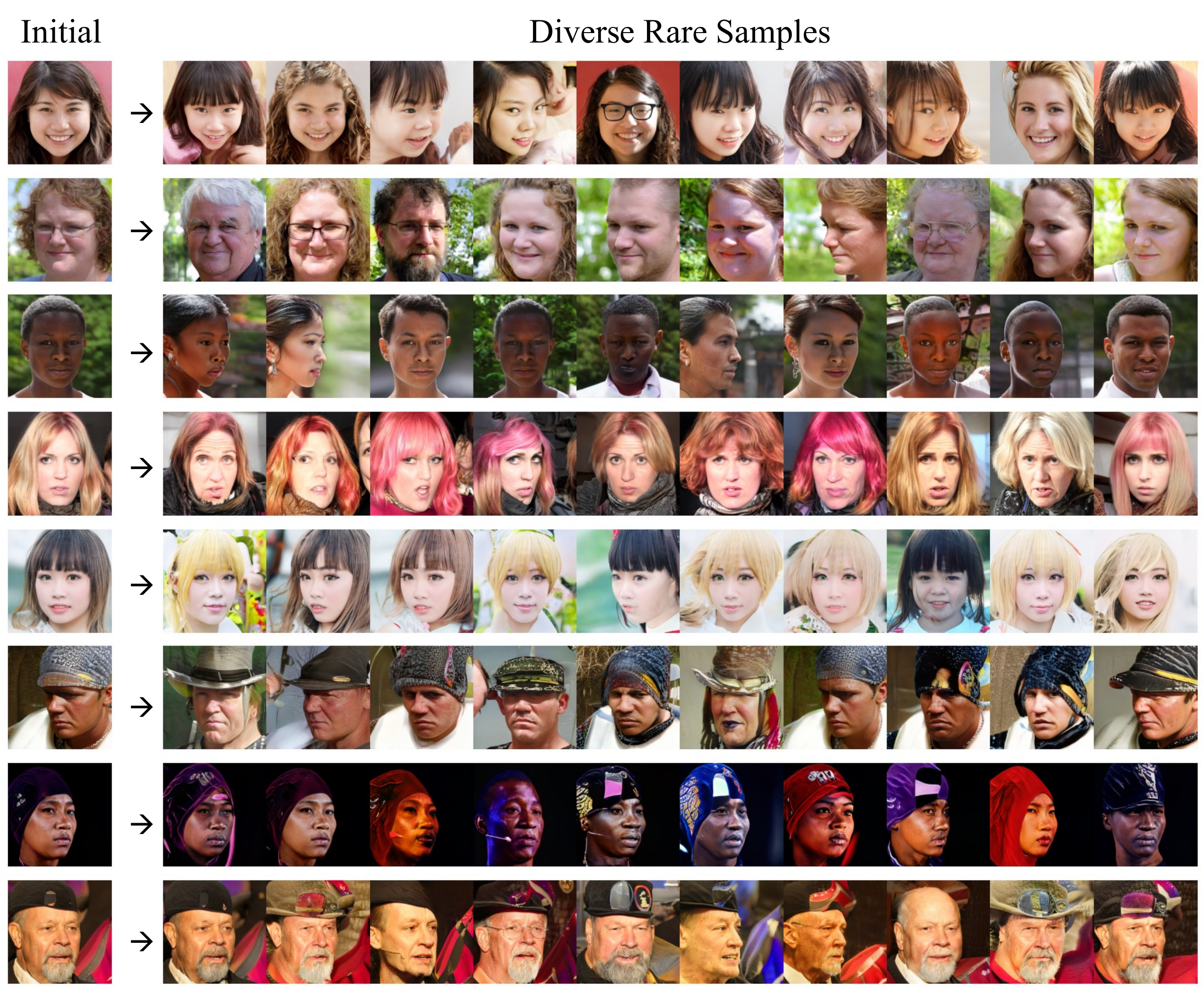}
    \caption{Examples of diverse rare samples generated by our method using FFHQ-StyleGAN2.}
    \label{fig:11}
\end{figure}

\begin{figure*}[!t]
    \centering    \includegraphics[width=0.99\linewidth]{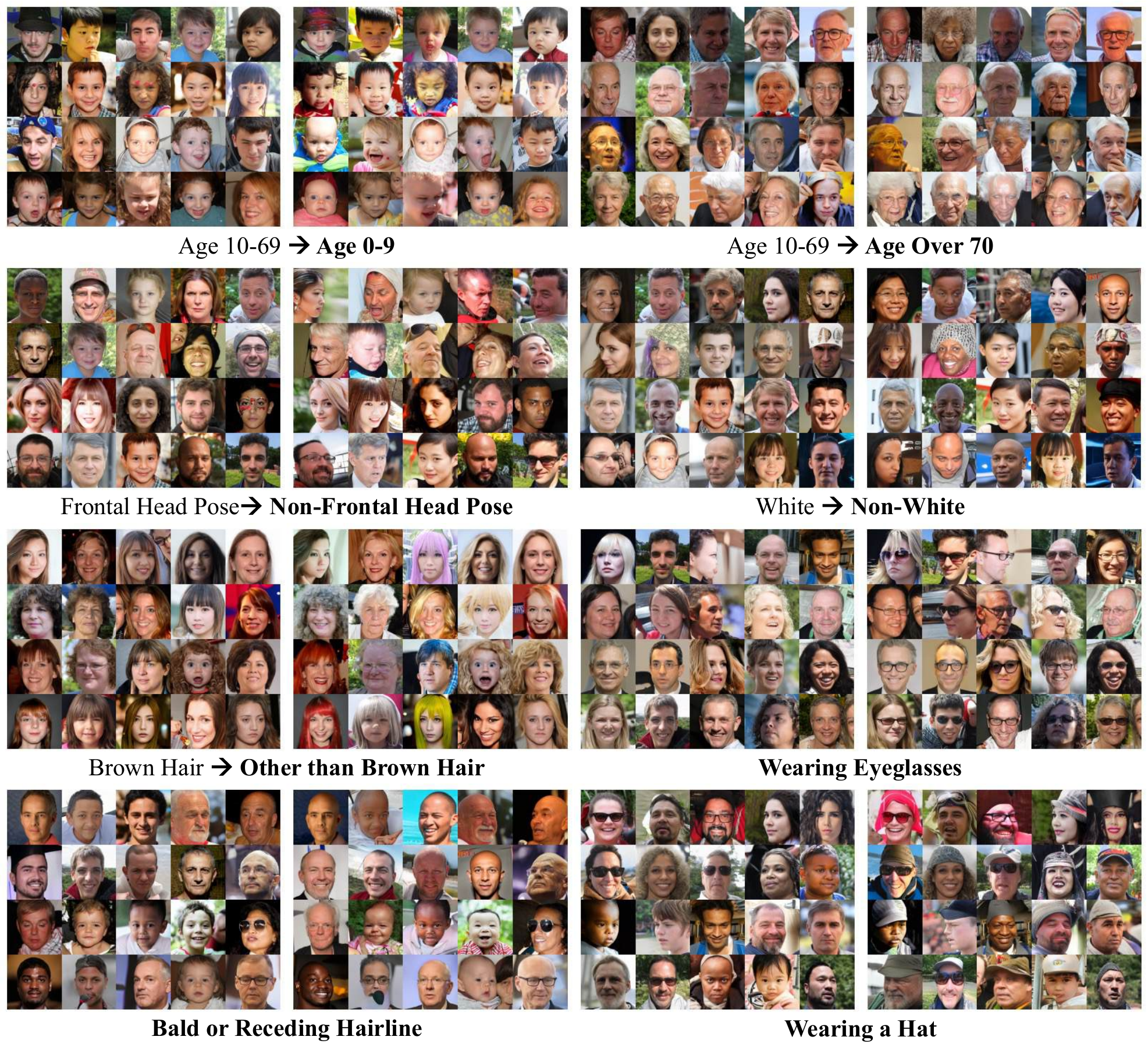}
    \caption{Additional rare samples generated by our method using FFHQ-StyleGAN2. In each row, the first and third columns serve as references for the second and fourth columns, respectively. The changed or generated attributes are listed below the figures. Rare attributes are highlighted in bold.}
    \label{fig:12}
\end{figure*}

\begin{figure*}
    \centering
    \includegraphics[width=0.99\linewidth]{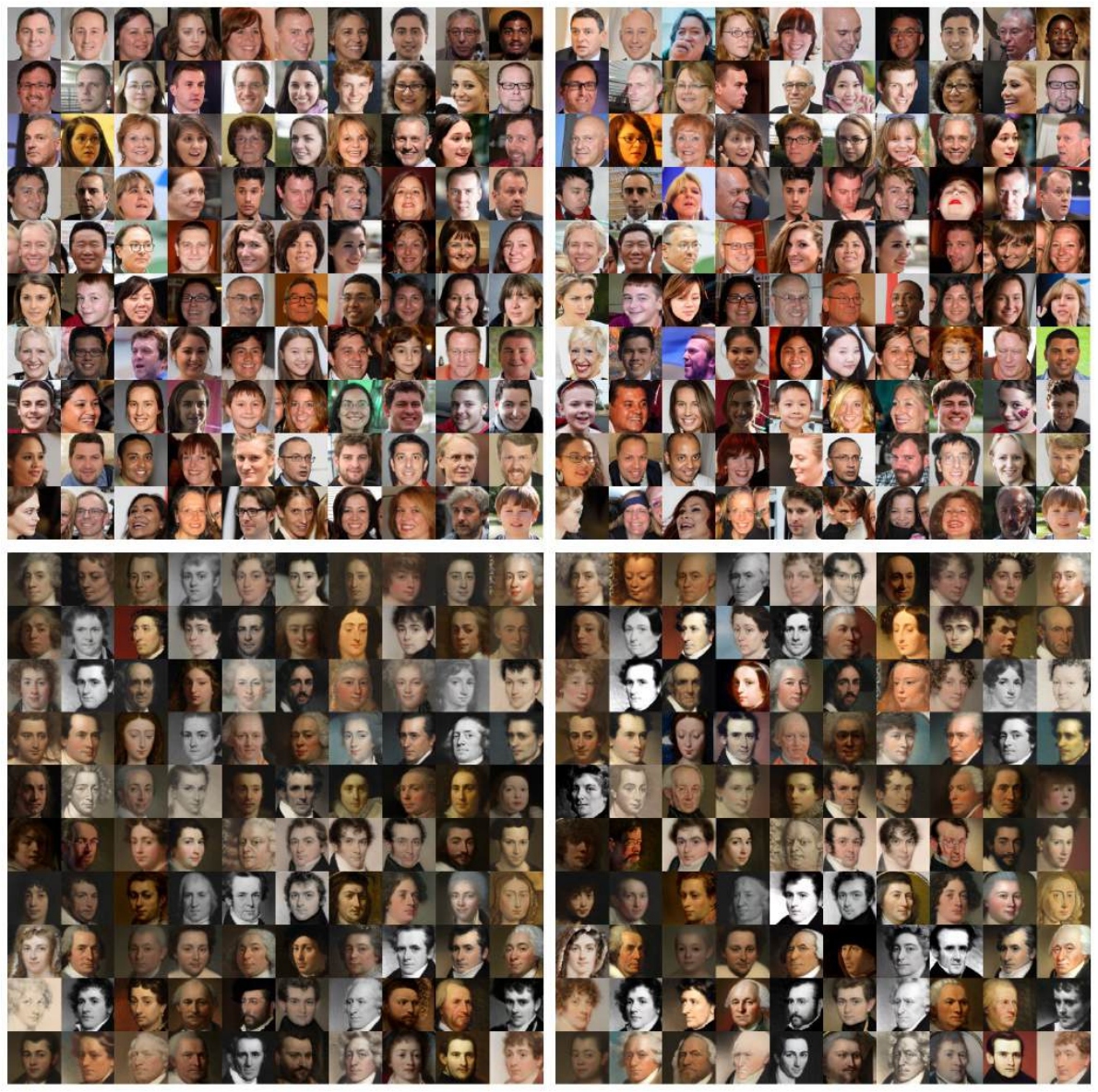}
    \caption{High-likelihood references (top 100) and their optimized rare images generated by our method. Top: FFHQ-StyleGAN2. Bottom: MetFaces-StyleGAN2-ADA. Left: References with a truncation value of $\psi=1.0$. Right: Optimized images.}
    \label{fig:16}
\end{figure*}

We also provide additional qualitative results in Fig.~\ref{fig:11},~\ref{fig:12}, and~\ref{fig:16} (top). Rare attributes include extreme ages, non-frontal head poses, non-white races, hair colors other than brown, eyeglasses, hairless features, and hats as shown in Fig.~\ref{fig:12}. From the top of Fig.~\ref{fig:16}, our method successfully changes the high-likelihood references into rarer ones.

\paragraph{About Artifacts} Some images in the results show low fidelity and contain undesirable artifacts. Although we use a real $k$-NN manifold and a penalizing boundary to prevent out-of-distribution samples, such artifacts are inevitable due to overestimated regions by the manifold assumption. Our objective function pushes samples toward low-density or even out-of-distribution regions. If these regions are included in the assumed real manifold, they may be selected as the best images by our algorithm. Improving the objective function or best sample selection process could help mitigate these issues and enhance the results.

\begin{figure*}
    \centering
    \includegraphics[width=0.99\linewidth]{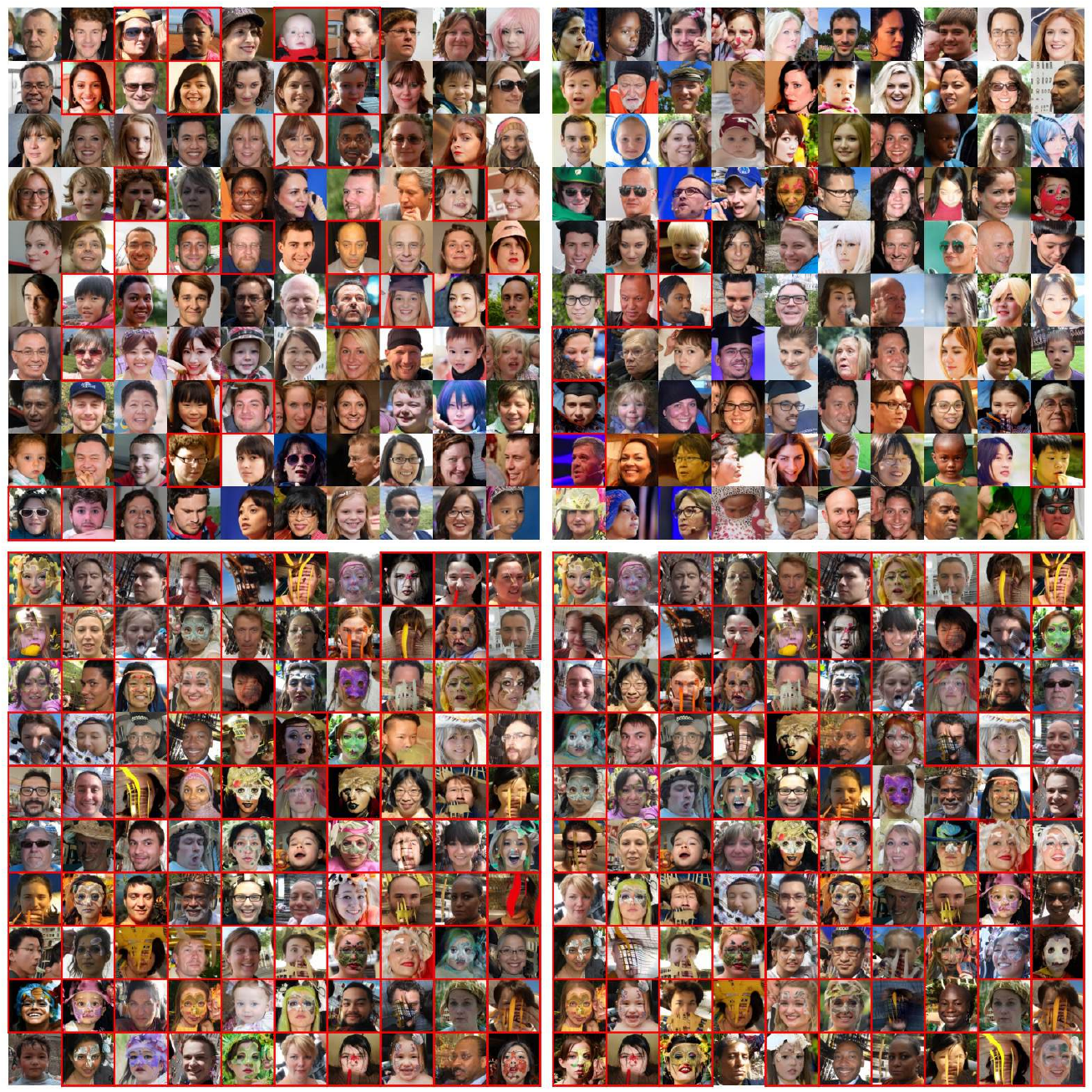}
    \caption{Comparative qualitative results: FFHQ-StyleGAN2 with a truncation value of $\psi=1.0$ (top-left), our method (top-right), and Polarity sampling with a truncation value of $\rho=1.0$ (bottom-left) and $\rho=5.0$ (bottom-right). Red boxes indicate out-of-manifold samples.}
    \label{fig:13}
\end{figure*}

\paragraph{Comparative Qualitative Results} In Fig.~\ref{fig:13}, 100 random samples from different methods are visualized. The red boxes represent out-of-manifold samples from the real $k$-NN manifold. Although both samples from our method and Polarity sampling show rare attributes that rarely represented in the baseline, most samples in the results of Polarity sampling include huge artifacts on the face, which make the samples be detected as out-of-manifold samples. We will discuss more about Polarity sampling in Section G.

\section{Additional Results for Section 4.2}
\subsection{Categorization of Dog Classes}
There are 120 classes of dogs in the ImageNet dataset, and we construct eight high-level groups from those. We use ChatGPT-4 \cite{openai2023gpt4} for classification and description for each category, and the result is shown in Table~\ref{tab:16}.

\setlength{\tabcolsep}{3.2mm}
\begin{table*}[!t]
\centering
\begin{tabular}{lrrrlrrr}
\hline
\textbf{LFWA Attribute}
 &
  \textbf{MetF.} &
  \textbf{Reference} &
  \textbf{Ours} &
   \textbf{LFWA Attribute}
  &\textbf{MetF.} &
  \textbf{Reference} &
  \textbf{Ours(\%)} \\ \hline
\textit{\textbf{Bald}}           &0  &0  &\textbf{0.14} & \textit{\textbf{BigLips}}        &3.01  &2.42  &\textbf{2.75}  \\
\textit{\textbf{RosyCheeks}}     &0  &0  &0  & \textit{\textbf{RecedingHairline}}  &3.31  &\textbf{2.62}  &2.50  \\
\textit{\textbf{Eyeglasses}}     &0.07  &0.10  &\textbf{0.33}  & \textit{\textbf{ArchedEyebrows}} &3.61  &4.44  &\textbf{5.32}  \\
\textit{\textbf{WearingNecklace}}&0.15  &\textbf{0.20}  &0.04  & \textit{\textbf{BlackHair}}      &4.51  &\textbf{4.54}  &4.34  \\
\textit{\textbf{Blurry}}         &0.15  &0.20  &\textbf{0.43}  & \textit{\textbf{OvalFace}}          &4.66  &\textbf{6.76}  &4.46  \\
\textit{\textbf{PointyNose}}     &0.22  &0.60  &\textbf{1.14}  & \textit{\textbf{WearingLipstick}}   &4.81  &3.43  &\textbf{4.86}  \\
\textit{\textbf{Mustache}}       &0.37  &0.10  &\textbf{0.16} & \textit{\textbf{Sideburns}}         &5.64  &5.35  &\textbf{6.36}  \\
\textit{\textbf{HeavyMakeup}}    &0.75  &0.30  &\textbf{0.89} & \textit{\textbf{5o'ClockShadow}} &6.40  &9.49  &\textbf{9.53}  \\
\textit{\textbf{WearingNecktie}} &0.82  &\textbf{0.70}  &0.60 &\textit{\textbf{MouthSlightlyOpen}} &8.88  &9.49  &\textbf{9.82}\\
\textit{\textbf{PaleSkin}}          &0.82  &0.90  &\textbf{2.08} &\textit{\textbf{Attractive}}     &9.18  &12.72  &\textbf{15.31} \\
\textit{\textbf{BlondHair}}      &1.50  &2.12  &\textbf{2.37} &\textit{\textbf{BushyEyebrows}}  &9.56  &\textbf{10.40}  &7.44\\
\textit{\textbf{HighCheekbones}} &1.65  &\textbf{1.51}  &0.79 & \textit{\textbf{Chubby}}         &10.24  &\textbf{9.89}  &8.99  \\
\textit{\textbf{StraightHair}}      &1.73  &2.02  &\textbf{2.23} & \textit{\textbf{BrownHair}}      &11.74  &\textbf{10.70}  &8.47  \\
\textit{\textbf{Smiling}}        &1.80  &1.01  &\textbf{1.06} & \textit{\textbf{WearingHat}}        &12.34  &9.89  &\textbf{12.18}  \\
\textit{\textbf{GrayHair}}       &1.80  &2.52  &\textbf{3.94} & \textit{\textbf{WavyHair}}          &25.75  &\textbf{28.48}  &21.93  \\
\textit{\textbf{Goatee}}         &1.95  &\textbf{2.82}  &2.08 &\textit{\textbf{BagsUnderEyes}}  &30.72  &\textbf{29.39}  &26.96\\
\textit{\textbf{NarrowEyes}}        &2.03  &\textbf{2.92}  &2.48 &\textit{\textbf{BigNose}}        &40.51  &\textbf{38.48}  &37.45\\
\textit{\textbf{WearingEarrings}}   &2.40  &1.61  &\textbf{2.08} & \textit{\textbf{Male}}              &70.33  &70.00 &\textbf{73.43}  \\
\textit{\textbf{DoubleChin}}     &2.48  &\textbf{2.62}  &2.37 & \textit{\textbf{Young}}             &76.50  &\textbf{80.10}  &74.16  \\
\textit{\textbf{Bangs}}          &2.48  &1.61  &\textbf{2.65} &\textit{\textbf{NoBeard}}           &85.09  &\textbf{86.46}  &84.22  \\
\hline
\end{tabular}
\caption{Percentage of LFWA attributes predicted by FaceXFormer. MetF. refers to the MetFaces dataset. 1,000 reference images are generated from MetFaces-StyleGAN2-ADA with a truncation value of $\psi=1.0$, and the optimized images generated by our method are derived from the initial latent vectors of these 1,000 references with $N=5$. Sorted in ascending order of MetF.(\%).}
\label{tab:17}
\end{table*}

\setlength{\tabcolsep}{2mm}
\begin{table}[t]
\centering
\begin{tabular}{ll|rrr}
\hline
\multicolumn{2}{l|}{\textbf{ImageNet Class}} & \textbf{Real} & \textbf{Reference} & \textbf{Ours(\%)} \\ 
\hline
\textbf{Cat} & \textbf{Tabby} & 56.76 & \textbf{58.00} & 55.10   \\
& \textbf{Siamese} & 15.33 & \textbf{16.80} & 15.87   \\
& \textbf{Persian} & 10.55 & 8.70 & \textbf{11.14}  \\
& \textbf{Egyptian} & 9.35 & \textbf{11.60} & 9.24    \\
& \textbf{Tiger Cat}& 4.71 & 2.80 & \textbf{5.26}   \\ 
\hline
\end{tabular}
\caption{Percentage of cat-related breeds in ImageNet classes. 1,000 reference images are generated by AFHQ Cat-StyleGAN2-ADA with a truncation value of $\psi=1.0$, and the optimized images generated by our method are derived from the initial latent vectors of these 1,000 references with $N=5$. The results are sorted in descending order of Real\%.}
\label{tab:12}
\end{table}

\subsection{Quantitative Results}
We provide the full version of Table~\ref{tab:4} for AFHQ Cat dataset and the generated cat face images, in Table~\ref{tab:12}. 
The percentage of Egyptian cats decreases compared to the reference samples despite not being major classes. This decrease occurs because this class in the reference samples are diversified to other classes during optimization.

The full version of Table~\ref{tab:5} is in Table~\ref{tab:17}. The percentages are calculated for the only face-detected cases. The percentages of the undetected cases are 0.59\%, 1.00\%, and 4.15\% for the MetFaces dataset, the references, and the optimized images, respectively. Compared to the FFHQ dataset, the MetFaces dataset has very low percentages of most of the LFWA attributes, where 26 attributes among the 40 attributes have a percentage lower than 5\%. For the ten most rare attributes in the MetFaces dataset, \textit{Bald, RosyCheeks, Eyeglasses, WearingNecklace, Blurry, PointyNose, Mustache, HeavyMakeup, PaleSkin, WearingNecktie}, seven attributes show an increased percentage in our method compared to the references. For the remaining three attributes, \textit{RosyCheeks} has zero percentage in all cases, and the percentages of \textit{WearingNecklace} and \textit{WearingNecktie} rather decreased in our method, which are disappeared when getting diverse. Other than those top ten attributes, our method increased the percentage of blond and gray hair, eyeglasses, earrings, hats, etc. Additionally, we found the over-trust issue of \textit{Male} in the FaceXFormer LFWA attributes classifier in the MetFaces dataset.

\subsection{Qualitative Results}
\begin{figure*}
    \centering
    \includegraphics[width=0.99\linewidth]{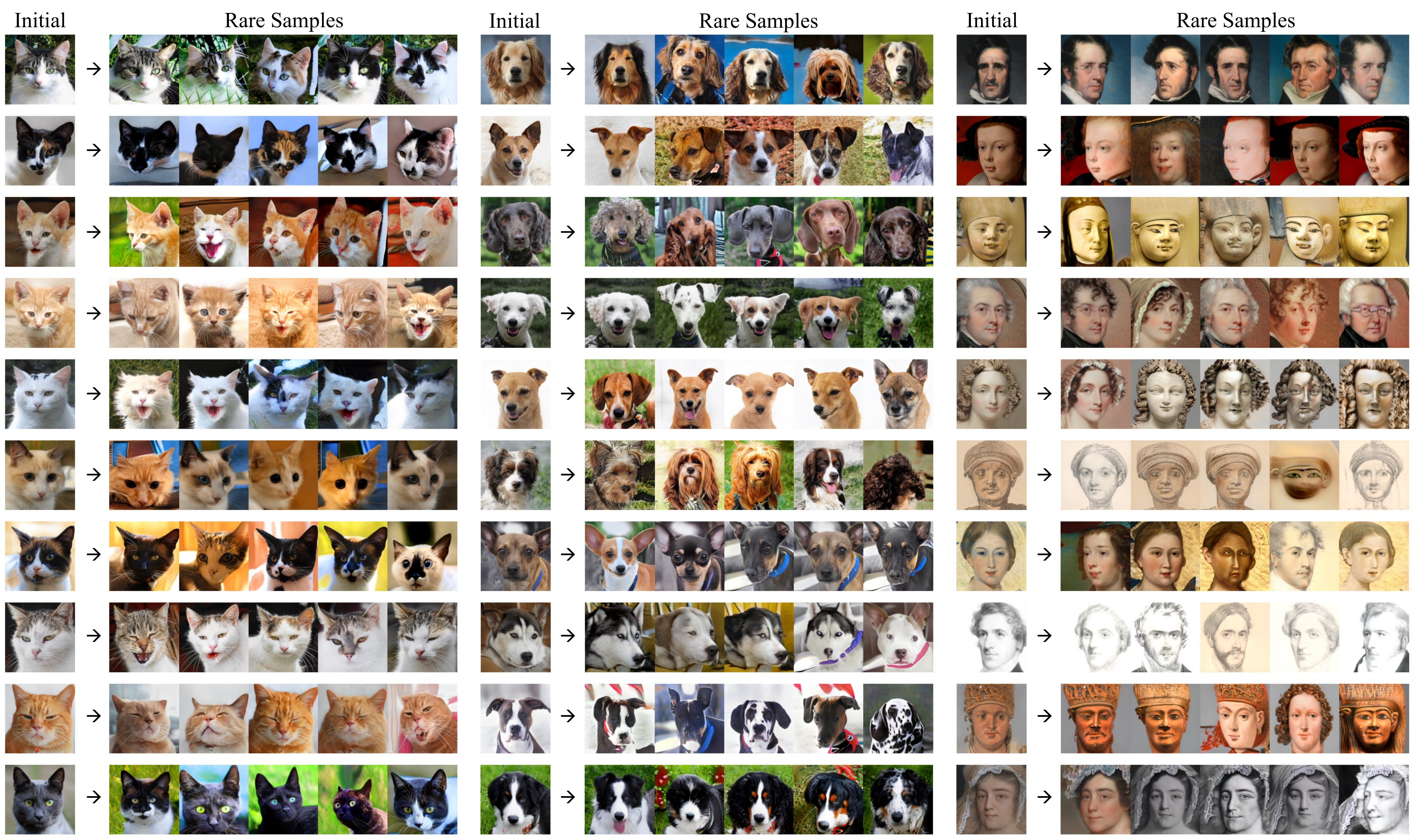}
    \caption{Examples of diverse rare samples generated by our method using AFHQ and MetFaces with StyleGAN2-ADA.}
    \label{fig:14}
\end{figure*}

\begin{figure*}
    \centering
    \includegraphics[width=0.99\linewidth]{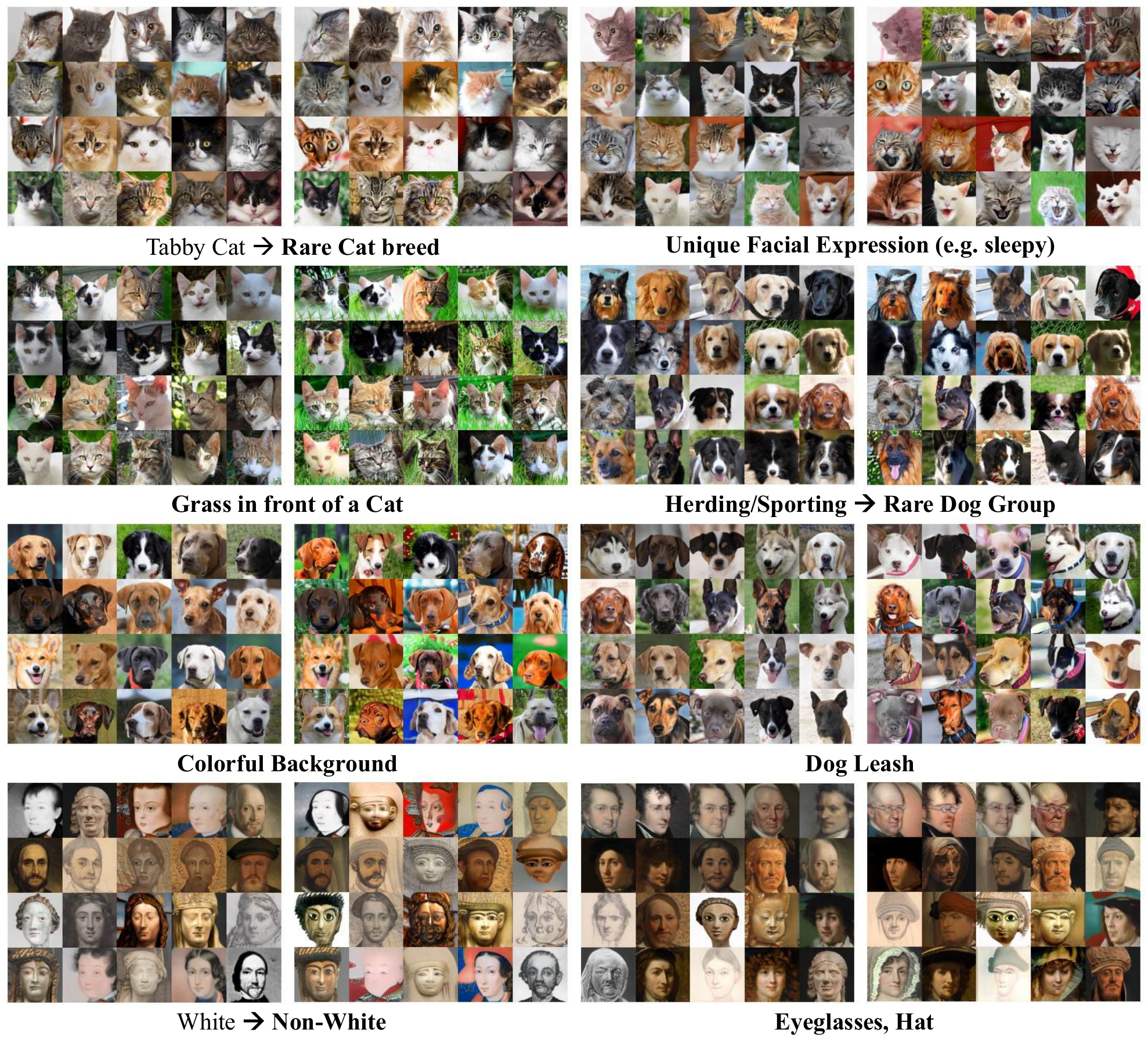}
    \caption{Additional rare samples generated by our method using AFHQ and MetFaces with StyleGAN2-ADA. In each row, the first and third columns serve as references for the second and fourth columns, respectively. The changed or generated attributes are listed below the figures. Rare attributes are highlighted in bold.}
    \label{fig:15}
\end{figure*}

\begin{figure*}
    \centering
    \includegraphics[width=0.99\linewidth]{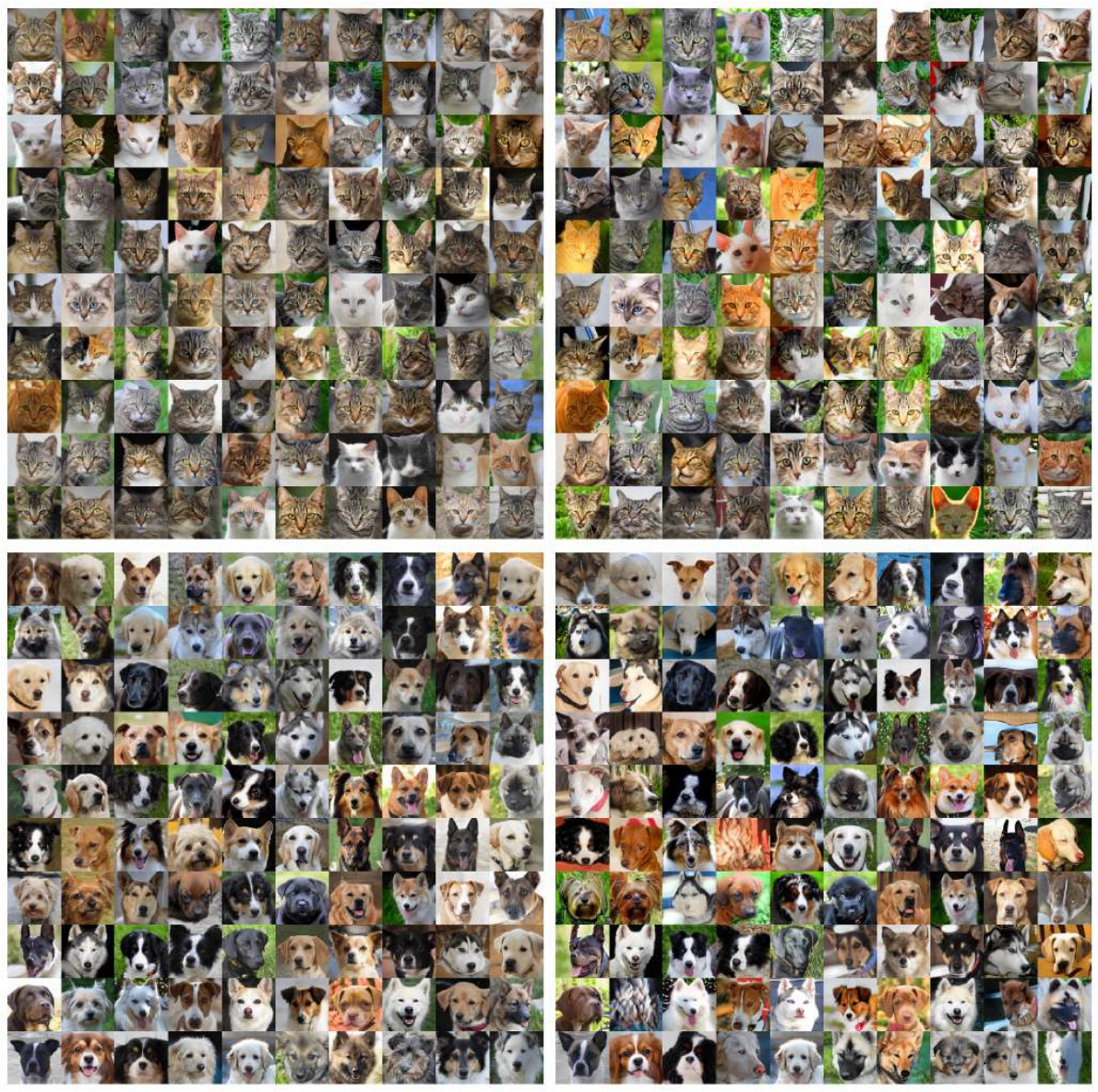}
    \caption{High-likelihood references (top 100) and their optimized rare images generated by our method. Top: AFHQ Cat-StyleGAN2-ADA. Bottom: AFHQ Dog-StyleGAN2-ADA. Left: References with a truncation value of $\psi=1.0$. Right: Optimized images.}
    \label{fig:17}
\end{figure*}

We provide additional qualitative results in Fig.~\ref{fig:14},~\ref{fig:15},~\ref{fig:17}, and the bottom of the Fig.~\ref{fig:16}.

\section{Experimental Setting and Additional Results for Section 4.4}\label{appx:manifold}

We fit the UMAP model for the real feature vectors to reduce the dimensionality from 4096 to 2. To draw a local region of the feature space with probability density estimated by the NF model, we sample the grid points from the dimensionality-reduced two-dimensional plane by the UMAP and transform them into the feature space by the inverse mapping function, followed by computing the $\log p(\mathbf{x})$ by the NF model. We interpolate them by the thin plate spline kernel $r^2 \times \log(r)$, a spline-based smoothing kernel that interpolates polynomials piecewisely. Technically, we use a Python API \texttt{UMAP}~\cite{mcinnes2018umap} and \texttt{scipy.interpolate.RBFInterpolator}~\cite{2020SciPy-NMeth} for interpolation. We compute the $k$-NN balls with $k=3$ on the transformed space to visualize them into the heatmap plausibly. With the Euclidean distance metric, we set the number of neighboring sample points to 15 to reduce the dimension to two as a hyperparameter setting for the UMAP. We set the smoothing parameter to zero and the degree of the kernel polynomial to first order. The other hyperparameter options follow the default settings. In the experiment, we fix the random state at 42.

\begin{figure*}[t]
    \centering
    \includegraphics[width=0.9\linewidth]{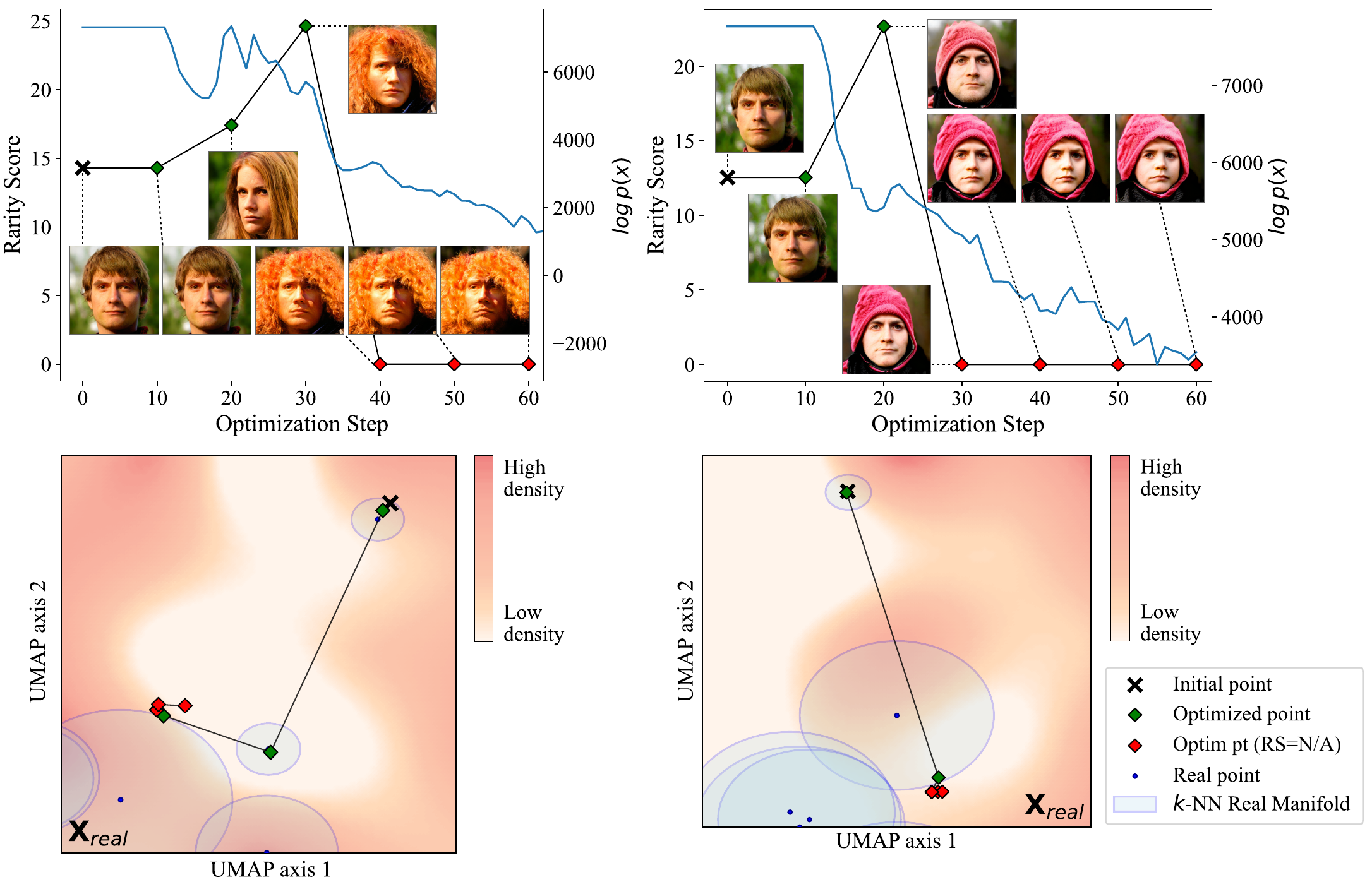}
    \caption{Examples of the optimization paths with a real $k$-NN manifold and a heatmap of likelihoods estimated by the normalizing flow. Top: The black line represents the optimization path, with each marker indicating a rarity score at every ten steps. The blue line represents the log-likelihood estimated by the NF model. Bottom: The balls indicate the nearby real $k$-NN manifold.}
    \label{fig:18}
\end{figure*}

We provide additional results in Fig.~\ref{fig:18}, demonstrating the relationship between rarity scores and NF-estimated likelihood. The optimization path is directed towards low density or larger real $k$-NN balls. However, our objective function allows the optimization path to continuously trail the real feature manifold, even the out-of-manifold area undefined by $k$-NN balls. In the cases in Fig.~\ref{fig:18}, rare attributes are obtained such as curly orange hair, a pink turban, and a non-frontal head pose.

\section{Experimental Setting and Additional Results for Polarity Sampling}
\subsection{Experimental Setting}
For Polarity sampling in Section 4.1, we utilize the pre-calculated latent vectors and the Jacobian matrix of the StyleGAN2-config f generator provided by the authors in \cite{humayun2022polarity}. Note that the latent seeds are different from our seeds. The corresponding GitHub repository is available at: \url{https://github.com/AhmedImtiazPrio/magnet-polarity}. Following the default settings, the singular value matrix is truncated to the top 30 values, and sampling is performed without replacement.

While we used the same number of generated samples from Polarity sampling in all statistical measures, in practice, this process involves generating a substantial number of initial samples and calculating their Jacobian matrices before resampling to achieve the desired sample numbers. Note that obtaining a large number of rare samples requires a pre-sampled set and pre-calculated Jacobians. Alternatively, in an online sampling setting, a very large number of samplings would be needed.
\setlength{\tabcolsep}{2mm}
\begin{table}[!ht]
\centering
\begin{tabular}{l|rrrrr}
\hline
\textbf{\,Polarity $\rho$} & \textbf{RS$\uparrow$} & \textbf{Prec.$\uparrow$} & \textbf{Rec.$\uparrow$} & \textbf{LPIPS$\uparrow$} & \textbf{FID$\downarrow$} \\ \hline
\textbf{\,0.1} &20.80  &0.60  &0.63  &0.74  & 7.69 \\ 
\textbf{\,0.3} &23.35  &0.45  &\textbf{0.71}  &0.74  &23.12  \\ 
\textbf{\,0.5} &24.17  &0.41  &\textbf{0.71}  &0.75  &29.48  \\ 
\textbf{\,0.8} &24.66  &0.39  &\textbf{0.71}  &0.75  &32.43  \\ 
\textbf{\,1.0} & 24.71 & 0.39 & 0.70 & 0.75 & 33.28 \\ 
\textbf{\,5.0} & \textbf{24.83} & 0.38 & \textbf{0.71} & 0.75 & 34.11 \\ 
\hline
\textbf{\,Baseline} & 18.88 & 0.69 & 0.56 & 0.73 & \textbf{4.17} \\ 
\textbf{\,Ours} & 23.50 & \textbf{0.92} & 0.65 & \textbf{0.76} & 7.38 \\ \hline
\end{tabular}
\caption{Quantitative evaluation of Polarity sampling for FFHQ and StyleGAN2 with varying $\rho$. RS refers to the rarity score.}
\label{tab:13}
\end{table}

\subsection{Different $\rho$'s}
We provide additional experimental results with different $\rho$'s in Table~\ref{tab:13}. From $\rho\geq0.5$, the Polarity sampling shows higher rarity scores compared to our method. However, all the listed results show significantly lower precision compared to the reference and our method.

\setlength{\tabcolsep}{1mm}
\begin{table}
\centering
\begin{tabular}{l|rrrrr}
\hline
\textbf{\begin{tabular}[c]{@{}l@{}}\,Polarity $\rho$\\ w/ Replacement\end{tabular}} & \textbf{RS$\uparrow$} & \textbf{Prec.$\uparrow$} & \textbf{Rec.$\uparrow$} & \textbf{LPIPS$\uparrow$} & \textbf{FID$\downarrow$} \\ \hline
\textbf{\,1.0} &27.21 &0.16 &0.53 &0.60 &219.74  \\ 
\textbf{\,5.0} &26.38 &0.0004 &0.21 &0.002 &299.73  \\ 
\hline
\end{tabular}
\caption{Quantitative evaluation of Polarity sampling with replacement for FFHQ and StyleGAN2. RS refers to the rarity score.}
\label{tab:14}
\end{table}

\setlength{\tabcolsep}{0.8mm}
\begin{table}[t]
\centering
\begin{tabular}{l|rrrrr}
\hline
\textbf{\begin{tabular}[c]{@{}l@{}}\,Polarity $\rho$\\ w/ Online Rejection\end{tabular}} & \textbf{RS$\uparrow$} & \textbf{Prec.$\uparrow$} & \textbf{Rec.$\uparrow$} & \textbf{LPIPS$\uparrow$} & \textbf{FID$\downarrow$} \\ \hline
\textbf{\,1.0} & 23.77 &1.00 & 0.58 & 0.74 & 20.54  \\ 
\textbf{\,5.0} &23.84 &1.00 &0.58 &0.74 & 21.28 \\ 
\hline
\end{tabular}
\caption{Quantitative evaluation of Polarity sampling for FFHQ and StyleGAN2, using online rejection sampling to prevent out-of-manifold samples. RS denotes the rarity score.}
\label{tab:15}
\end{table}

\begin{figure}
    \centering
    \includegraphics[width=0.99\linewidth]{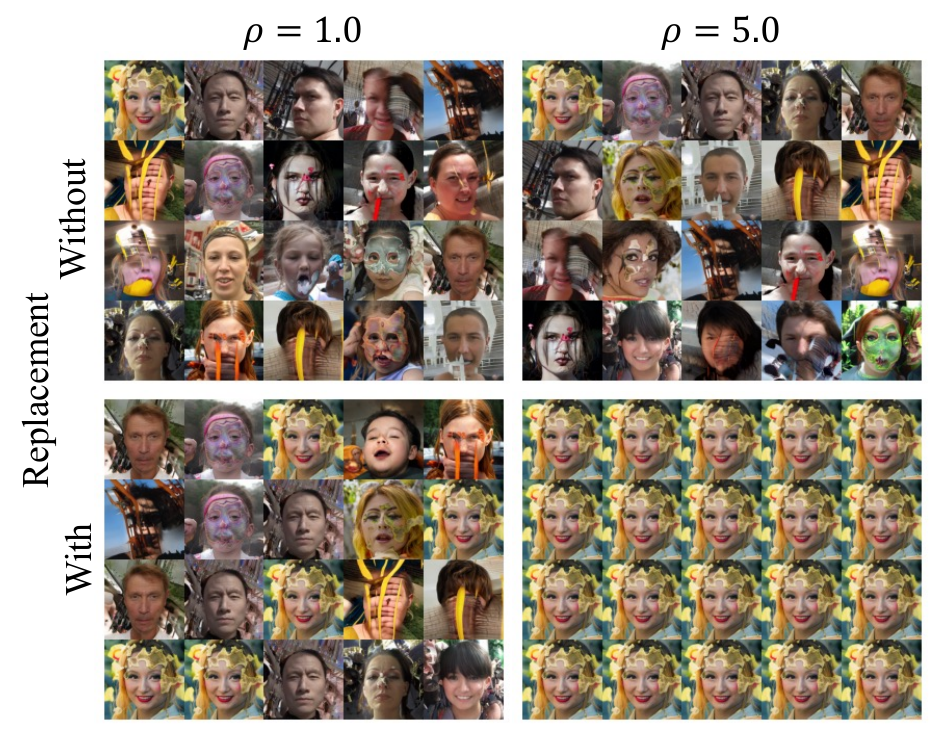}
    \caption{Qualitative results of Polarity sampling. Top: Sampling without replacement. Bottom: Sampling with replacement. Left: $\rho=1.0$. Right: $\rho=5.0$. Each set contains 20 randomly generated samples.}
    \label{fig:19}
\end{figure}

\subsection{Sampling with Replacement}
For practical purposes of maintaining diversity, the replacement parameter has been set to false during Polarity sampling. However, sampling without replacement can introduce bias into the results. Notably, in the earlier work by the same authors, MaGNET sampling \cite{humayun2021magnet}, which provides the theoretical foundation for Polarity sampling, the resampling procedure was defined with replacement. We conduct Polarity sampling with replacement, and the results are shown in Table~\ref{tab:14} and Fig.~\ref{fig:19}. With replacement, certain out-of-manifold samples are resampled very frequently, reducing the diversity of results and limiting the opportunity to sample in-distribution rare samples.

\subsection{Online Rejection Sampling}
From Table~\ref{tab:3}, Polarity sampling with a positive $\rho$ can obtain rare samples, but at the cost of a very high percentage of out-of-manifold samples. In this section, to investigate the true capability of Polarity sampling, we utilized online rejection sampling. This method collects the same number of samples sequentially while rejecting the out-of-manifold samples. As a result, all the collected samples are within the real $k$-NN manifold, indicating a precision of 1.

\begin{figure*}
    \centering
    \includegraphics[width=0.99\textwidth]{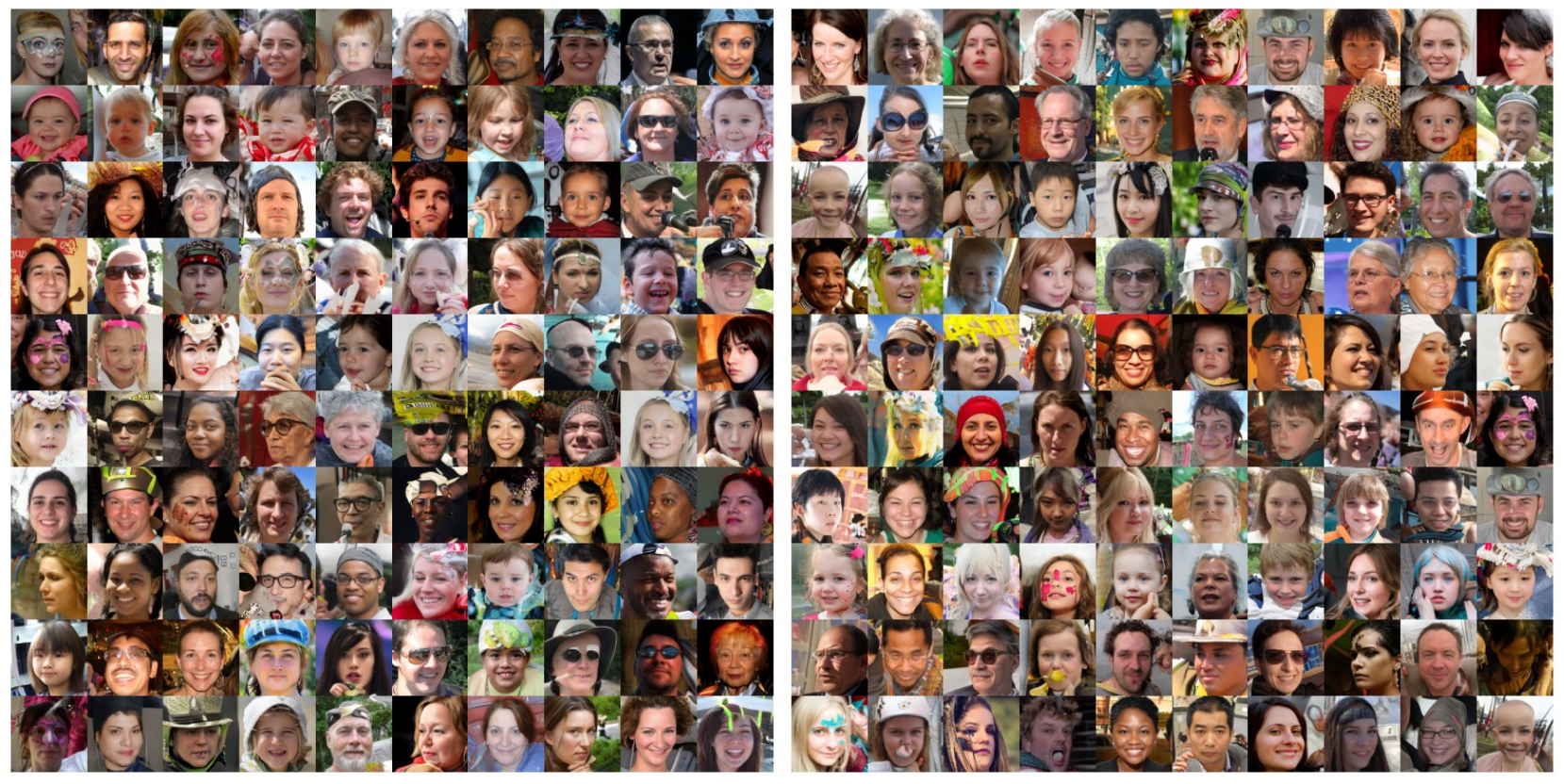}
    \caption{Qualitative results of Polarity sampling with $\rho=1.0$ (left) and $\rho=5.0$ (right), using online rejection sampling to prevent out-of-manifold samples. All samples are within the real $k$-NN manifold constructed from the FFHQ dataset.}
    \label{fig:20}
\end{figure*}

With $\rho=1.0$, 22,285 samples are generated to collect 10,000 valid samples. Similarly, with $\rho=5.0$, 22,376 samples are generated to collect 10,000 valid samples, requiring more than twice as many samples. We recalculate the rarity score, recall, LPIPS, and FID scores for these samples and presented the results in Table~\ref{tab:15}. We provide the qualitative results in Fig.~\ref{fig:20}.

Compared to the original Polarity sampling, replacing out-of-manifold samples with in-manifold samples results in a decrease in the $k$-NND of the samples which previously had out-of-manifold neighbors. This reduction in $k$-NND within the fake manifold leads to a decrease in recall. This also implies that the similarity between samples increases, which leads to a decrease in the LPIPS score.
If the sampling is performed with replacement, this issue would be more significant, since a few rare samples with very high resampling weights would be selected frequently. Compared to our method, Polarity sampling with online rejection achieves similar average rarity and fidelity. However, our method generates a greater diversity of rare samples.

\setlength{\tabcolsep}{2mm}
\begin{table*}
\centering
\begin{tabular}{l|ll}
\hline
\textbf{Dog Group}    & \textbf{ImageNet Class Number}                                                                                                                  & \textbf{Description}                                                                                                       \\ \hline
\textbf{Toy}          & \begin{tabular}[c]{@{}l@{}}151, 152, 153, 154, 155, 157, 158, 171, 185, 186, \\ 187, 200, 201, 252, 254, 259, 262, 265\end{tabular}      &Small dogs, often with short, round faces.                                     \\ \hline
\textbf{Hound}        & 159, 160, 169, 170, 172, 173, 176, 177, 253                                                                                              & \begin{tabular}[c]{@{}l@{}}Dogs with long faces, often having long snouts and \\large ears.\end{tabular}               \\ \hline
\textbf{Scent Hound}  & 161, 162, 163, 164, 165, 166, 167, 168, 174, 175                                                                                         & \begin{tabular}[c]{@{}l@{}}Broad-faced dogs with typically droopy ears.\end{tabular}                                   \\ \hline
\textbf{Terrier}      & \begin{tabular}[c]{@{}l@{}}179, 180, 181, 182, 183, 184, 188, 189, 190, 191, \\ 192, 193, 194, 196, 199, 202, 203\end{tabular}           & \begin{tabular}[c]{@{}l@{}}Dogs with short, sturdy faces and pronounced, \\strong snouts.\end{tabular}                    \\ \hline
\textbf{Sporting}     & \begin{tabular}[c]{@{}l@{}}156, 178, 205, 206, 207, 208, 209, 210, 211, 212, \\ 213, 214, 215, 216, 217, 218, 219, 220, 221\end{tabular} & \begin{tabular}[c]{@{}l@{}}Medium-sized dogs, usually with balanced features.\end{tabular}                              \\ \hline
\textbf{Non-Sporting} & 195, 204, 223, 245, 251, 260, 261, 266, 267, 268                                                                                         & \begin{tabular}[c]{@{}l@{}}A diverse group of dogs with a variety of face shapes \\and body types.\end{tabular}        \\ \hline
\textbf{Herding}      & \begin{tabular}[c]{@{}l@{}}197, 198, 224, 225, 226, 227, 228, 229, 230, 231, \\ 232, 233, 235, 240, 241, 263, 264\end{tabular}           & \begin{tabular}[c]{@{}l@{}}Dog breeds for herding and managing livestock.\end{tabular}                                  \\ \hline
\textbf{Working}      & \begin{tabular}[c]{@{}l@{}}222, 234, 236, 237, 238, 239, 242, 243, 244, 246, \\ 247, 248, 249, 250, 255, 256, 257, 258\end{tabular}      & \begin{tabular}[c]{@{}l@{}}Large, powerful dog breeds for tasks like \\guarding, pulling, and rescue work.\end{tabular} \\ \hline
\end{tabular}
\caption{Dog groups categorized from 120 dog classes in ImageNet dataset.}
\label{tab:16}
\end{table*}

\end{document}